\documentclass[journal]{IEEEtran}

\usepackage[T1]{fontenc}
\usepackage{amsmath}
\usepackage{amsfonts}
\usepackage{amssymb}
\usepackage{graphicx}
\usepackage{algorithm}
\usepackage{algorithmicx}
\usepackage{algpseudocode}
\usepackage{booktabs}
\usepackage{array}
\usepackage{multirow}
\usepackage[colorlinks]{hyperref}
\usepackage{cleveref}
\usepackage{xspace}
\usepackage{xcolor}
\usepackage{makecell}
\usepackage{colortbl}
\usepackage{subfigure}
\usepackage{CJK}
\usepackage{color}
\usepackage{mathrsfs}
\usepackage{url}

\newcommand{\ie}{\emph{i.e.},\ }

\newcommand{\et}{\emph{et al.}\ }

\newcommand{\bert}{\emph{BERT}\xspace}
\newcommand{\crname}{\emph{ClueReader}\xspace}
\newcommand{\bilstm}{\emph{Bi-LSTM}\xspace}

\newcommand{\BiDAF}{\emph{BiDAF}\xspace}
\newcommand{\BERT}{\emph{BERT}\xspace}
\newcommand{\Longformer}{\emph{Longformer}\xspace}

\newcommand{\GCNfulname}{\emph{Graph Convolutional Networks}\xspace}
\newcommand{\CogQA}{\emph{CogQA}\xspace}
\newcommand{\EntityGCN}{\emph{Entity-GCN}\xspace}
\newcommand{\ELMo}{\emph{ELMo}\xspace}
\newcommand{\BAG}{\emph{BAG}\xspace}
\newcommand{\Glove}{\emph{Glove}\xspace}
\newcommand{\PathbasedGCN}{\emph{Path-based GCN}\xspace}
\newcommand{\HDE}{\emph{HDE}\xspace}
\newcommand{\coattention}{\emph{co-attention}\xspace}
\newcommand{\selfpooling}{\emph{self-attention}\xspace}
\newcommand{\glove}{\emph{GloVe}\xspace}
\newcommand{\pytorch}{\emph{PyTorch}\xspace}
\newcommand{\pyg}{\emph{PyTorch Geometric}\xspace}
\newcommand{\nx}{\emph{NetworkX}\xspace}

\newcommand{\Wikidata}{\textsc{Wikidata}\xspace}
\newcommand{\Wikipedia}{\textsc{Wikipedia}\xspace}

\newcommand{\WikiHop}{\textsc{WikiHop}\xspace}
\newcommand{\MedHop}{\textsc{MedHop}\xspace}
\newcommand{\MEDLINE}{\textsc{MEDLINE}\xspace}
\newcommand{\DrugBank}{\textsc{DrugBank}\xspace}

\newcommand{\qangaroo}{\textsc{QAngaroo}\xspace}
\newcommand{\nltk}{\textsc{NLTK}\xspace}
\newcommand{\spacy}{\textsc{spaCy}\xspace}
\newcommand{\hotpotqa}{\textsc{HotpotQA}\xspace}
\DeclareMathAlphabet{\mathscr}{OT1}{pzc}{m}{it}
\newcommand{\edit}[1]{\textcolor{black}{#1}}

\begin{document}

\title{ClueReader: Heterogeneous Graph Attention Network for Multi-hop Machine Reading Comprehension}

\author{
Peng~Gao,
Feng~Gao,
Peng~Wang,
Jian-Cheng~Ni,
Fei~Wang,
and~Hamido~Fujita

\thanks{P. Gao and P. Wang are with the School of Cyber Science and Engineering, Qufu Normal University, Qufu, Shandong 273165, China.}
\thanks{F. Gao is with the School of Computer Science and Technology, East China Normal University, Shanghai 200062, China, and also with the Shanghai Institute of AI for Education, Shanghai 200062, China.}
\thanks{J.-C. Ni is with the Network and Information Center, Qufu Normal University, Qufu, Shandong 273165, China.}
\thanks{F. Wang is with School of Electronics and Information Engineering, Harbin Institute of Technology, Shenzhen, Guangdong 518055, China.}
\thanks{H. Fujita is with the Faculty of Information Technology, HUTECH University, Ho Chi Minh City 70000, Vietnam, the Andalusian Research Institute in Data Science and Computational Intelligence (DaSCI), University of Granada, Granada 18011, Spain, and also with the Research and Regional Cooperation Division, Iwate Prefectural University, Takizawa, Iwate 020-0611, Japan.}
}

\maketitle

\begin{abstract}

Multi-hop machine reading comprehension is a challenging task in natural language processing as it requires more reasoning ability across multiple documents. Spectral models based on graph convolutional networks have shown good inferring abilities and lead to competitive results. However, the analysis and reasoning of some are inconsistent with those of humans. Inspired by the concept of \emph{grandmother cells} in cognitive neuroscience, we propose a heterogeneous graph attention network model named \crname to imitate the \emph{grandmother cell} concept. The model is designed to assemble the semantic features in multi-level representations and automatically concentrate or alleviate information for reasoning through the attention mechanism. The name \crname is a metaphor for the pattern of the model: it regards the subjects of queries as the starting points of clues, takes the reasoning entities as bridge points, considers the latent candidate entities as \emph{grandmother cells}, and the clues end up in candidate entities. The proposed model enables the visualization of the reasoning graph, making it possible to analyze the importance of edges connecting entities and the selectivity in the mention and candidate nodes, which is easier to comprehend empirically. Evaluations on the open-domain multi-hop reading dataset \WikiHop and drug-drug interaction dataset \MedHop proved the validity of \crname and showed the feasibility of its application of the model in the molecular biology domain.

\end{abstract}

\IEEEpeerreviewmaketitle
\section{Introduction}\label{se:intro}

Machine reading comprehension (MRC) is one of the most attractive and long-standing tasks in natural language processing (NLP).
Compared with single-paragraph MRC, multi-hop MRC is more challenging since multiple confusing answer candidates are contained in different passages \cite{wang-etal-2018-multi-passage,dai2021multiple}.
Models designed for this task are supposed to have abilities to reasonably traverse multiple passages and discover reasoning clues following given questions.
For complex multi-hop MRC tasks, more understandable, reliable, and analyzable methodologies are required to improve reading performance.

A better understanding of biological brains could play a vital role in building artificial intelligent systems \cite{HASSABIS2017245}.
Previous cognitive research in reading can be of benefit to challenging multi-hop MRC tasks.
The concept of \emph{grandmother cells} can be traced back to a $1969$ academic lecture given by the neuroscientist Jerome Lettvin \cite{page2000connectionist}, and was later defined by the physiologist Horace Barlow as cells in the brain that respond specifically to a single familiar person or object. In experiments on primates, researchers discovered individual neurons that responded specifically to a specific person, image, or concept after differentiation \cite{dehaene2009reading}. A study of a patient with epilepsy found a neuron in the patient's anterior temporal lobe that responded specifically to the Hollywood star Jennifer Aniston \cite{quiroga2005invariant}. Any form of stimulation related to Aniston, whether it be a color photograph, a close-up of her face, a cartoon portrait, or even just seeing her name written on paper, could and would only stimulate that neuron to produce an excited signal. As research into the concept of \emph{grandmother cells}, the underlying mechanism of their response became clearer. The signal output from a single \emph{grandmother cell} in response to specific stimuli actually stems from the coordinated calculation of a large-scale neural network behind \emph{grandmother cells} \cite{dehaene2009reading}. It suggests that a single neuron can respond to only one out of thousands of stimulation, which is somehow intuitively similar to reading and inference in multi-hop MRC:
\begin{itemize}
    \item \textbf{Selectivity.} The \textit{grandmother cells} concept organizes the neurons in a hierarchical ``sparse'' coding scheme. It activates some specific neurons to respond to stimulation, similar to the manner in which we store reasoning evidence maps (neurons) in our minds during reading and recall-related evidence maps to reason the answer with a question (stimulation) constrained.
    \item \textbf{Specificity.} The concept implies that brains contain \textit{grandmother} neurons that are so specialized and dedicated to a specific object, which is similar to a particular MRC question resulting in a specific answer among multiple reading passages and their complex reasoning evidence.
    \item \textbf{Class character.} Amazing selectivity is captured in \emph{grandmother cells}. However, it results from computation by much larger networks and the collective operations of many functionally different low-level cells, similar to human multi-hop reading in which evidence is usually gathered from different levels as much as possible and the final answer is decided in some candidate endpoints.
\end{itemize}

To imitate \emph{grandmother cells} in multi-hop MRC, the reading evidence is supposed to be organized as level-classified neurons and the selections must be performed in response to specific question stimulation.
As for multi-hop MRC tasks, the hops between two entities could be connected as node pairs and gradually constructed into a reasoning evidence graph taking all related entities as nodes.
This reasoning evidence graph is intuitively represented as a graph structure, which can be empirically considered to contain the implicit reasoning chains from the start of the question to the end of the answer nodes (entities).
We generally recall considerable related evidence as a node, whatever form it is (such as a paragraph, a short sentence, or a phrase) to meet the class character, and we coordinate their inter-relationship before obtaining the results.

Graph neural networks (GNNs) inspire us to posit that operating on graphs and manipulating the structured knowledge can support relational reasoning \cite{DBLP:journals/corr/abs-1806-01261,journal2} in a sophisticated and flexible pattern, similar to the implementation of \textit{grandmother cells} regarding the cells as nodes in the graph and collecting evidence in multi-classified aspects of node representations.
Further, spatial graph attention networks (GATs) perform the selectivity in the reasoning evidence graph in the manner of \emph{grandmother cells} using attention mechanisms.
This work has the following main contributions:
\begin{enumerate}
    \item In order to construct a more reasonable graph, \crname draws inspiration from the concept of \textit{grandmother cells} in the brain during information cognition, in which cells in the brain only output specific entities. This leads to the creation of heterogeneous graph attention networks with multiple types of nodes.
    \item By taking the subject of queries as the starting point, potential reasoning entities in multiple documents as bridge points, and mention entities consistent with candidate answers as end points, the proposed \crname is a heuristic way of constructing MRC chains.
    \item Before outputting predicted answers, \crname innovatively visualizes the internal state of the heterogeneous graph attention network, providing intuitive quantitative data displays for analyzing the effectiveness, rationality, and explainability.
\end{enumerate}

The remainder of the article is organized as follows.
Section \ref{sec:2} describes the work related to multi-hop MRC, and Section \ref{sec:3} proposes the \crname that imitates \emph{grandmother cells} for multi-hop MRC.
Experimental evaluations are conducted in Section \ref{sec:4}, and conclusions are summarized in Section \ref{sec:5}.

\section{Related Work}\label{sec:2}

\subsection{Sequential Reading Models for Multi-hop MRC}

Sequential reading models were first used for single-passage MRC tasks, and most of them are based on recurrent neural networks (RNNs) or their variants.
When the attention mechanism was introduced into NLP tasks, their performance significantly improved \cite{DBLP:conf/iclr/SeoKFH17, DBLP:conf/acl/CuiCWWLH17,journal1, li2023modeling}.
In the initial benchmarks of the \qangaroo \cite{qan}, a dataset for multi-hop MRC, the milestone model \emph{Bi-Directional Attention Flow} (\BiDAF) \cite{DBLP:conf/iclr/SeoKFH17} was first applied to evaluate its performance in the multi-hop MRC task.
It represented the context at different levels and used a bi-directional attention flow mechanism to obtain query-aware context representation and was then used for predictions.

Some studies \cite{bert, DBLP:conf/nips/VaswaniSPUJGKP17, DBLP:journals/corr/abs-1907-11692, DBLP:journals/corr/abs-2004-05150} argued that independent attention mechanisms, \ie \emph{Bidirectional Encoder Representations from Transformers} (\BERT) \cite{bert}-style models, applied on sequential contexts can outperform former RNN-based approaches in various NLP downstream tasks including MRC.
When the sequential approaches were applied to multi-hop MRC tasks, however, they suffered from the challenge that the super-long contexts --- to adapt the design of the sequential requirement, multiple passages are concatenated into one passage --- resulted in dramatically increased calculation and time consumption.
A long-sequence architecture, \Longformer \cite{DBLP:journals/corr/abs-2004-05150}, overcomes the self-attention restriction and allows the length of sequences to be increased from 512 to 4,096 and then concatenates all the passages into a long sequential context for reading.
The \Longformer modified the question answering (QA) methodology proposed in \BERT \cite{bert}: the long sequential context consisted of a question, candidates, and passages, which were separated by special tags that were applied to the linear layers to output the predictions, while still having enough memory for first 4,096 length sequence.

Although the approaches above are effective, \cite{DBLP:journals/corr/abs-2202-07206} indicate that model reasoning is not robust enough. We consider that there are still two main challenges that should be further addressed:
(1) With the expansion of the problem scale and the reasoning complexity, the token-limited problem may appear again eventually.
For instance, a full-wiki setting task in \hotpotqa requires models to predict answers from the scope of the entire \Wikipedia, which is a dataset for diverse and explainable multi-hop question answering.
It is difficult to imagine how a huge search space is built based on a large amount of text.
(2) Some models which simply concatenate text to long contexts lack logical relationships, which is unconvincing in terms of their reasoning.
Thus, the approaches based on GNNs were proposed to improve the scalability and explainability in multi-hop MRC.

\begin{figure*}[]
    \centering
    \includegraphics[width=\textwidth]{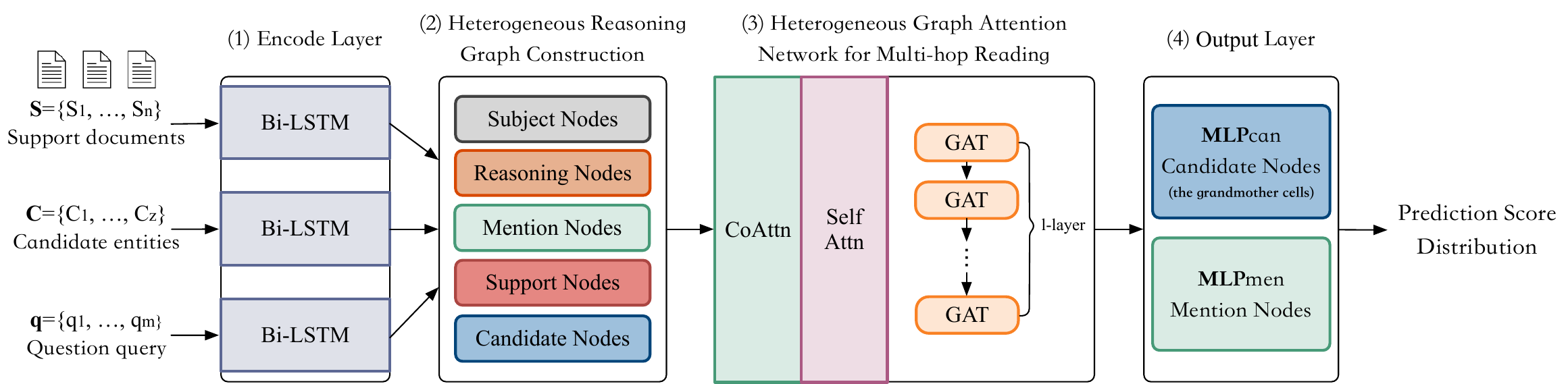}
    \caption{Our proposed \crname: a heterogeneous graph attention network for multi-hop MRC. The detailed explanations of $S$, $C$, and $q$ are in task formalization (Section \ref{subsec:task-formalization}).
    $S$, $C$, and $q$ are encoded in three independent \bilstm (Section \ref{subsec:encoder-layer}).
    Following the graph construction strategies in Section \ref{subsec:graph}, the outputs of three encoders are applied to \emph{Co-attention} and \emph{Self-attention} to initialize the reasoning graph features, which is explained in Section \ref{subsec:coatten-selfatten}.
    Then the topology information and node features are passed into the GAT layer.
    A much larger network computation behind \emph{grandmother cells} is performed in GAT Layer, and n-hops message passing is calculated in n parameter shared layers which are represented in Section \ref{subsec:graph-update}.
    Finally, \emph{grandmother cells} selectivity is combined in Section \ref{subsec:output-layer}, outputting the final predicted answer.
    }
    \label{fig:model-architecture}
\end{figure*}

\subsection{Graph Neural Networks for Multi-hop MRC}

Reasoning about explicitly structured data, in particular, graphs has arisen at the intersection of deep learning and structured approaches \cite{DBLP:journals/corr/abs-1806-01261}.
As the representative graph methodology, \GCNfulname (GCNs) \cite{gcn,zhang2023graph} are widely applied in multi-hop MRC approaches.
\emph{Cognitive Graph QA} (\CogQA) \cite{DBLP:conf/acl/DingZCYT19} was founded on the dual process theory \cite{evans1984heuristic, sloman1996empirical}, and it divides the multi-hop reading process into two stages: the implicit extraction (System I) based on \BERT and the explicit reasoning (System II) established in GCNs.
System I extracts the answer candidates and useful next-hop entities from passages for the cognitive graph construction, then System II updates entity representations and predict the final answer in the GCN message passing way.
In this procedure, the selected passages are not put in the system at once.
As a result, \CogQA keeps its scalability in the face of the massive scope of reading materials.

\EntityGCN \cite{DBLP:conf/naacl/CaoAT19} extracts all the text spans matching the candidates as nodes and obtains their representations from the contextualized \ELMo \cite{DBLP:conf/naacl/PetersNIGCLZ18} word embeddings, then passes them to the GCN module for reasoning.
Based on \EntityGCN, \emph{Bi-directional Attention Entity Graph Convolutional Network} (\BAG) \cite{DBLP:conf/naacl/CaoFT19} added \Glove word embeddings and two manual features, named-entity recognition and part-of-speech tags, to reflect the semantic properties of tokens.
On account of the full usage of the question contextual information, it applies the bi-directional attention mechanism, both node2query, and query2node, to obtain query-aware node representations in the reasoning graph for better predictions.
\PathbasedGCN \cite{DBLP:conf/ijcai/TangSMXYL20} introduces more related entities in the graph than the nodes merely matching the candidates to enhance the performance of the model.
\emph{Heterogeneous Document-Entity} (\HDE) model \cite{hde} introduces the heterogeneous nodes into GCNs, which contain different granularity levels of information.
Additionally, \emph{Keywords-Aware Dynamic Graph Neural Network} (\emph{KA-DGN}) \cite{JIA202225} was proposed and designed as a dynamic graph neural network to further tackle reading over multiple scattered text snippets.
Furthermore, Zhang \et \cite{YingZHANG20222021EDP7154} and Song \et \cite{10.1109/TKDE.2020.2982894} separately proposed knowledge-aware and evidence-aware GNN reading models, which integrate dependency relations or multiple pieces of evidence from multiple paragraphs.

However, the reading process of the above-mentioned approaches is still inexplicable, especially in GNNs, which stimulated our interest in the selectivity of this procedure.

\section{Methodology}\label{sec:3}

We introduce the design and implementation of the proposed model, \crname, which is shown in Figure~\ref{fig:model-architecture}.

\subsection{Task Formalization}\label{subsec:task-formalization}
A given query $q=(s, r, a^*)$ is in a triple form, where $s$ is the subject entity, $r$ is the query relation (\ie predication), and $q$ can be converted into sequential form $q=\{q_1, q_2, ..., q_m\}$, where $m$ is the number of tokens in the query $q$.
Then a set of candidates $C_q=\{c_1, c_2, ..., c_{\edit{z}}\}$ and a series of supporting documents $S_q{=\{s_1, s_2, ..., s_n\}}$ containing the candidates are also provided, where $z$ is the number of the given candidates, $n$ is the number of the given supporting documents, and the subscript $q$ means the two sets are constrained by the query $q$.
Moreover, $S_q$ is provided in a random order, and without $S_q$, the answer to the query $q$ could be multiple.
Our goal is to identify the single correct answer $a^*\in C_q$ by reading $S_q$.
\subsection{Encoding Layer}\label{subsec:encoder-layer}

We utilize the pre-trained \glove \cite{pennington2014glove} model to initialize word embeddings, and then employ \emph{Bidirectional Long Short-Term Memory} (\bilstm) \cite{lstm1997, DBLP:conf/acl/ZhouSTQLHX16} to encode sequence representations as:
\begin{equation}
  \begin{aligned}
    \left( \begin{array}{c}
             f_t \\
             i_t \\
             o_t
           \end{array}\right)&=\sigma(W_hh_{t-1}+W_ix_t)\\
    \tilde{c}_t &= \tanh(W_hh_{t-1}+W_ix_t)\\
    c_t &= f_tc_{t-1}+i_t\tilde{c}_t\\
    h_t &= o_t\tanh(c_t)
  \end{aligned}
\end{equation}

where the subscripts $t$ and $t-1$ denote the indexes of encoding time step; $W_i$ and $W_h$ are the hyperparameters of the input and the hidden layer; $i$, $f$, $o$, $\tilde{c}$, $h$ and $c$ respectively represent the input, forget, output, content, hidden and cell states; $x$ represents the word embedding; $\sigma$ and ${\rm tanh}$ are sigmoid activation and hyperbolic tangent activation, respectively.

We use $\overrightarrow{h}$ and $\overleftarrow{h}$ to denote the forward-pass (\ie the left-to-right) and the backward-pass (\ie the right-to-left) sequence representations encoded by \bilstm, respectively.
Then, the representation of the entire sequential context obtained from the encoding layer can be expressed as follows:

\begin{equation}
    h=[\overrightarrow{h}||\overleftarrow{h}]
\end{equation}
where the symbol $||$ denotes the concatenation of $\overrightarrow{h}$ and $\overleftarrow{h}$.
To encode the sequence representations of support documents $S$, candidates $C$, and query $q$, it is desirable to use three independent \bilstm. Their outputs are $H_s^i\in\mathbb{R}^{l_s^i\times d}$, $H_c^j\in\mathbb{R}^{l_c^j\times d}$ and $H_q\in\mathbb{R}^{l_q\times d}$, respectively, where $i$ and $j$ are the indexes of the documents and the candidates, $l$ is the sequence length, and $d$ is the output dimension of the representations.

\begin{figure*}[]
\centering
    \includegraphics[width=0.5\linewidth]{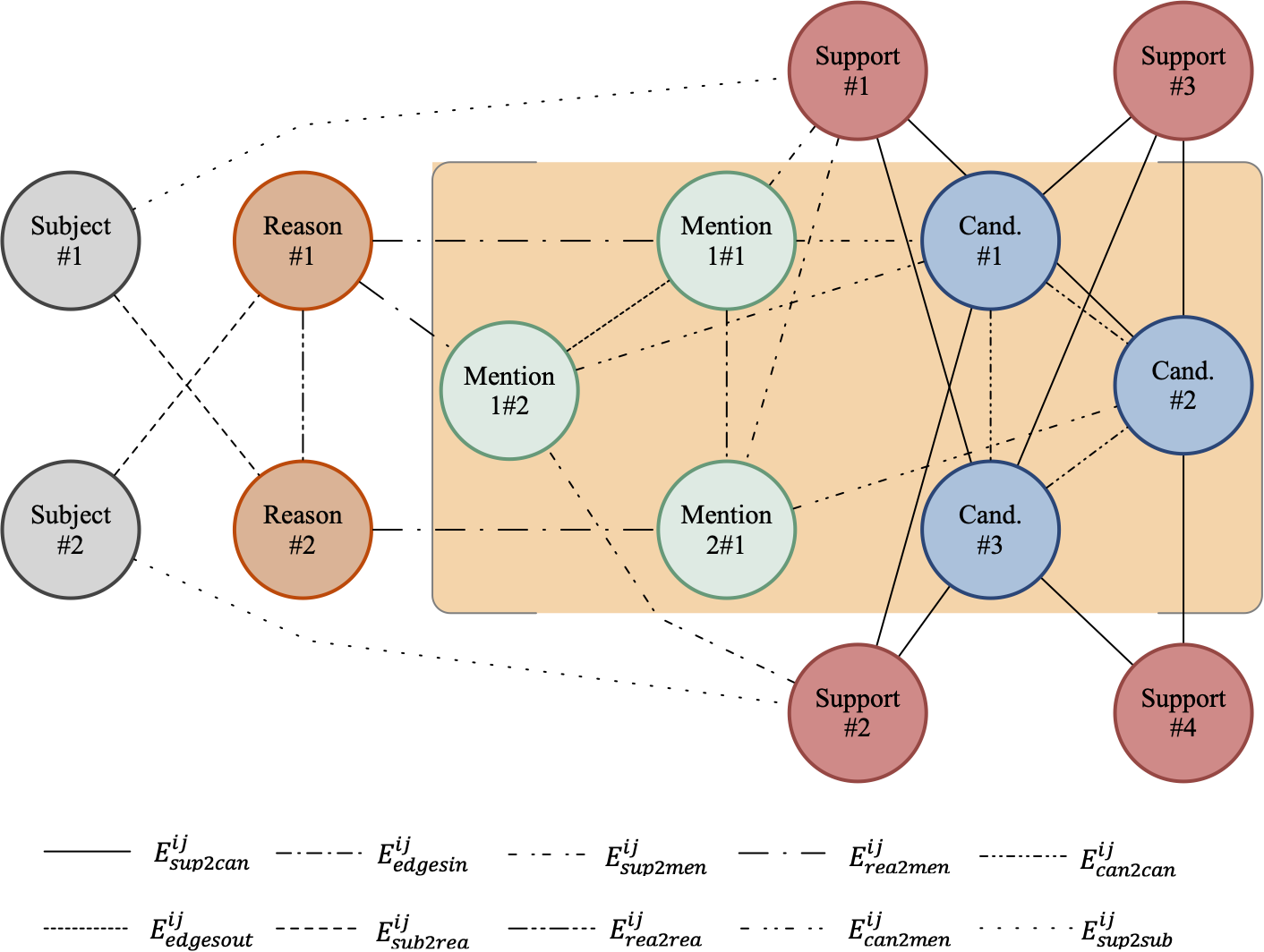}
\caption{Heterogeneous reasoning graph in \crname. Different nodes are filled in different colors, and the edges are distinguished by the types of lines. Subject nodes are gray, reasoning nodes are orange, mention nodes are green, support nodes are red, and candidate nodes are blue. The nodes in the light yellow square are all selected to input to the two \textbf{MLP} obtaining the prediction score distribution.}
    \label{fig:node-reasoning}
\end{figure*}

\subsection{Heterogeneous Reasoning Graph}\label{subsec:graph}

The concept of \emph{grandmother cells} reveals that the brains of monkeys, like those of humans, contain neurons that are so specialized they appear to be dedicated to a single person, image, or concept.
This amazing selectivity is uncovered in a single neuron, while it must result from computation by a much larger network \cite{dehaene2009reading}.
We heuristically consider that this procedure in multi-hop reading could be summarized as three steps:
\begin{enumerate}
  \item The query (or the question) locates the related neurons at a low level, which then stimulates higher-level neurons to trigger computation;
  \item The higher-level neurons begin to respond to increasingly broader portions of other neurons for reasoning and to avoid a broadcast storm, informative selectivity takes place in this step;
  \item At the top level, some independent neurons are responsible for the computations that occurred in step 2. We refer to these neurons as \emph{grandmother cells} and expect them to provide the appropriate results that correspond to the query.
\end{enumerate}

We attempt to imitate \emph{grandmother cells} in our reading procedure and present our reasoning graph as consistent as possible with the three steps mentioned above.
The heterogeneous reasoning graph $\mathscr{G}=\{\mathscr{V}, \mathscr{E}\}$, which is illustrated in Figure~\ref{fig:node-reasoning}, simulates a heuristic chain of comprehension that starts from the subject entity in query $q$ and goes through the reasoning entities in the supporting document set $S_q$, then through the mention entities in $S_q$ that are consistent with the candidate answer, and finally touches at the candidates in set $C_q$ (referred to as the \emph{grandmother cell}).

\subsubsection{Nodes Definition}

To construct the graph, we define five different types of nodes which are similar to neurons and ten kinds of edges among the nodes \cite{DBLP:conf/nips/VaswaniSPUJGKP17,DBLP:conf/naacl/CaoAT19}.

\begin{itemize}
    \item \textbf{Subject Nodes} --- As the form of query $q$, the subject entity $s$ is given in $q=(s, r, a^*)$.
    For example, the subject entity of the query sequence context \emph{Where is the basketball team that Mike DiNunno plays for based}? is certainly \emph{Mike DiNuuno}.
    We extract all the named entities that match with $s$ from documents and regard them as the subject nodes to open up the reading clues triggering further computations.
    The subject nodes are denoted as $\edit{\mathscr{V}}_{sub}$ and colored in gray in Figure~\ref{fig:node-reasoning}.
    \item \textbf{Reasoning Nodes} --- In light of the requirements of the multi-hop MRC, there are some gaps between the subject entities and candidates. To build bridges between the two and make the reasoning clues as complete as possible, we replenish those clues with the named recognition entities and nominal phrases from the documents containing the question subjects and answer candidates.
    The reasoning nodes are marked as $\edit{\mathscr{V}}_{rea}$ and colored in orange in Figure~\ref{fig:node-reasoning}.
    \item \textbf{Mention Nodes} --- A series of candidate entities are given in $C_q$, they may occur in multiple times within the document set $S_q$. As a result, we traverse the documents and extract the named entities corresponding to each candidate as mention nodes, serving as the soft endpoint of the reasoning chain. It should be noted that mention nodes will participate in the semi-supervised learning process and will be involved in the final answer prediction.
    The mention nodes are presented as $\edit{\mathscr{V}}_{men}$ and colored in green in Figure~\ref{fig:node-reasoning}.
    \item \textbf{Support Nodes} --- As described by \cite{dehaene2009reading}, we consider that multi-type representations may contribute to the reading process, thus the support documents containing the above nodes are introduced to $\mathscr{G}$ as support nodes, which are notated as $\edit{\mathscr{V}}_{\sup}$ and colored in red in Figure~\ref{fig:node-reasoning}.
    \item \textbf{Candidate Nodes} --- To imitate \emph{grandmother cells}, we consider candidate nodes as hard endpoints of the reasoning chain to gather relevant information from the heterogeneous reasoning graph. For the mention nodes $\mathscr{V}^q_{men}$ of a candidate answer $c_q$, when $\mathscr{V}_{men}^q\geq1$, candidate nodes are established as \emph{grandmother cells} to provide the final prediction.
    The candidate nodes are denoted as $\edit{\mathscr{V}}_{can}$ and colored in blue in Figure~\ref{fig:node-reasoning}.
\end{itemize}

\subsubsection{Edges Definition}\label{subsec:edge-def}
To learn the entity relationships between different nodes, we define 10 kinds of edges between nodes in heterogeneous reasoning graphs inspired by the literature \cite{tang-etal-2021-dureader,DBLP:conf/naacl/CaoAT19,DBLP:conf/naacl/CaoFT19}, as shown in Table~\ref{tab1}.

\begin{table*}[]
\caption{The definition of edges in the heterogeneous graph attention network \crname.\label{tab1}}
\centering
\begin{tabular}{l|p{16cm}}
\toprule
\textbf{Edges}	& \textbf{Definition}\\
\midrule
$\mathbf{\mathscr{E}_{sup2sub}}$ & If the support document $s_i$ contains the $j$-th subject node $\mathscr{v}_{sub}^j$, an undirected edge denoted as $\mathscr{e}_{sup2sub}^{ij}$ is established to connect the support node $\mathscr{v}_{sup}^i$ of $s_i$ and the subject node $\mathscr{v}_{sub}^j$.	\\\midrule
$\mathbf{\mathscr{E}_{sup2can}}$ & If the support document $s_i$ contains the $j$-th candidate node $\mathscr{v}_{can}^j$, an undirected edge denoted as $\mathscr{e}_{sup2can}^{ij}$ is established to connect the support node $\mathscr{v}_{sup}^i$ of $s_i$ and the candidate node $\mathscr{v}_{can}^j$. \\\midrule
$\mathbf{\mathscr{E}_{sup2men}}$ & If the support document $s_i$ contains the $j$-th mention node $\mathscr{v}_{men}^j$, an undirected edge denoted as $\mathscr{e}_{sup2men}^{ij}$ is established to connect the support node $\mathscr{v}_{sup}^i$ of $s_i$ and the mention node $\mathscr{v}_{men}^j$. \\\midrule
$\mathbf{\mathscr{E}_{can2men}}$ & If the $j$-th mention node $\mathscr{v}_{men}^j$ and the $i$-th candidate node $\mathscr{v}_{can}^i$ represent the same entity, an undirected edge denoted as $\mathscr{e}_{can2men}^{ij}$ is established to connect the two nodes. \\\midrule
$\mathbf{\mathscr{E}_{sub2rea}}$ & If the $i$-th subject node $\mathscr{v}_{sub}^{i}$ and the $j$-th reasoning node $\mathscr{v}_{rea}^{j}$ extracted from the same document, an undirected edge denoted as $\mathscr{e}_{sub2rea}^{ij}$ is established to connect the two nodes. \\\midrule
$\mathbf{\mathscr{E}_{rea2men}}$ & If the $i$-th reasoning node $\mathscr{v}_{rea}^{i}$ and the $j$-th mention node $\mathscr{v}_{men}^j$ extracted from the same document, an undirected edge denoted as $\mathscr{e}_{rea2men}^{ij}$ is established to connect the two nodes. \\\midrule
$\mathbf{\mathscr{E}_{can2can}}$ & All the mention nodes are fully connected using undirected edge $\mathscr{e}_{can2can}^{ij}$. \\\midrule
$\mathbf{\mathscr{E}_{edgesin}}$ & If two mention nodes $\mathscr{v}_{men}^i$ and $\mathscr{v}_{men}^j$ are extracted from the same document, the two nodes will be connected as $\mathscr{e}_{edgesin}^{ij}$. \\\midrule
$\mathbf{\mathscr{E}_{edgesout}}$ & If two mention nodes $\mathscr{v}_{men}^i$ and $\mathscr{v}_{men}^j$ are extracted from different documents represent the same entity, the two nodes will be connected as $\mathscr{e}_{edgesout}^{ij}$. \\\midrule
$\mathbf{\mathscr{E}_{rea2rea}}$ & If two reasoning nodes $\mathscr{v}_{rea}^{i}$ and $\mathscr{v}_{rea}^{j}$ are extracted from the same document or represent the same entity, the two nodes will be connected as $\mathscr{e}_{rea2rea}^{ij}$. \\
\bottomrule
\end{tabular}
\end{table*}

\subsubsection{Graph Construction}\label{subsec:graph-cons}

In the heterogeneous reasoning graph, the clue-reading chain can be represented by $\mathscr{V}_{sub}\leftrightarrow \mathscr{V}_{rea}\leftrightarrow \mathscr{V}_{men}\leftrightarrow \mathscr{V}_{can}$, whose edges are covered by $\mathscr{E}_{sub2rea}$, $\mathscr{E}_{rea2rea}$, $\mathscr{E}_{rea2men}$, and $\mathscr{E}_{can2men}$.
$\mathscr{E}_{edgesout}$ and $\mathscr{E}_{rea2rea}$ give the model the ability to transfer information across documents and edges in $\mathscr{E}_{sup2sub}$, $\mathscr{E}_{sup2can}$, and $\mathscr{E}_{sup2men}$ are responsible to supplement the multi-angle textual information from the documents.
Furthermore, the $\mathscr{E}_{can2men}$ could gather all the information of the mentioned nodes corresponding to the candidates and then pass their representations to the output layer to realize the imitation of \emph{grandmother cells}.

Specifically, this multi-hop MRC process of the clue-based reasoning starts with the subject node, connecting reasoning nodes from support documents, then connecting the mention nodes as soft endpoints of the clue chain, and finally connecting to the candidate nodes (\emph{grandmother cells}) as hard endpoints of the clue chain. For example, for the question, \emph{Which country is the location of the United Nations Headquarters?}, the answer candidate set includes \emph{China, France, UK, USA,} and \emph{Russia}. One correct and reasonable clue chain can be represented as \emph{Location of United Nations Headquarters} (subject node)$\leftrightarrow$\emph{Manhattan}$\leftrightarrow$\emph{New York City}$\leftrightarrow$\emph{New York State}$ \leftrightarrow$\emph{USA} (mention node)$\leftrightarrow$\emph{USA} (candidate node). In practice, multiple clue chains are included within the heterogeneous reasoning graph, and under the constraints of the query, the selection of soft and hard endpoints is required to output the final prediction.

\subsection{Heterogeneous Graph Attention Network for Multi-hop Reading}\label{subsec:coatten-selfatten}

\subsubsection{Query-aware Contextual Information}

Following \HDE \cite{hde}, we use the \coattention and \selfpooling mechanisms \cite{DBLP:conf/iclr/ZhongXKS19} to combine the query contextual information and documents. Moreover, it is applied to the other semantic representations that require reasoning consistent with the query.
To represent the query-aware support documents, it can be calculated as follows:

\begin{equation}
    \edit{\mathscr{A}}_{qs}^i=H_s^i\left(H_q\right)^\top\in\mathbb{R}^{l_s^i\times l_q}
\end{equation}
where $\edit{\mathscr{A}}_{qs}^i$ is the similarity matrix for two sequences, between the $i$-th support document $H_s^i\in\mathbb{R}^{l_s^i\times d}$ and query $H_q\in\mathbb{R}^{l_q\times d}$, and $d$ is the dimension of the context.
Then, the query-aware representation of support documents $S_{ca}$ is computed as follows:

\begin{equation}
    \edit{\mathscr{K}}_q={\rm softmax}\left(\edit{\mathscr{A}}_{qs}^\top\right)H_s\in\mathbb{R}^{l_q\times d}
\end{equation}

\begin{equation}
    \edit{\mathscr{K}}_s={\rm softmax}\left(\edit{\mathscr{A}}_{qs}\right)H_q\in\mathbb{R}^{l_s\times d}
\end{equation}

\begin{equation}
    \edit{\mathscr{D}}_s=\edit{{\rm BiLSTM}}\left({\rm softmax}\left(\edit{\mathscr{A}}_{qs}\right)\edit{\mathscr{K}}_q\right)\in\mathbb{R}^{l_s\times d}
\end{equation}

\begin{equation}    \edit{\mathscr{S}}_{ca}=\left[\edit{\mathscr{K}}_s||\edit{\mathscr{D}}_s\right]\in\mathbb{R}^{l_s\times2d}
\end{equation}

To project the sequence into a fixed dimension and output the representation $\edit{\mathscr{N}}_{sup}$ \edit{of $\mathscr{V}_{sup}$} for graph optimization, a \selfpooling is utilized to summarize the contextual information:

\begin{equation}
    \edit{\mathscr{j}}_s={\rm softmax}\left(\textbf{MLP}\left(\edit{\mathscr{S}}_{ca}\right)\right)\in\mathbb{R}^{l_s\times1}
\end{equation}

\begin{equation}
    \edit{\mathscr{N}}_{sup}=\edit{\mathscr{j}}_s^\top \edit{\mathscr{S}}_{ca}\in\mathbb{R}^{1\times2d}
\end{equation}

In addition to the query-aware support documents, the \coattention and \selfpooling are used to generate query-aware node representations from other sequential representations.

\subsubsection{Message Passing in the Heterogeneous Graph Attention Network}\label{subsec:graph-update}

We present messaging passing in the heterogeneous graph attention network for reading within multiple relations in diverse nodes. The input of this module is a graph \edit{$\mathscr{G}=\{\mathscr{V}, \mathscr{E}\}$} and node representations $\mathscr{N}=\{n_1,n_2,\ldots,n_r\}\in\mathbb{R}^{1\times2d}$, where $r$ is the number of nodes.
Initially, a shared weight matrix $\textbf{W}_{\edit{\mathscr{n}}}$ is applied to $\edit{\mathscr{N}}$, then the attention coefficients and nodes attention coefficients are computed as

\begin{equation}
    e_{ij}=\edit{\textbf{MLP}}\left(\textbf{W}_{\edit{{n}}}\edit{{n}_{i}}||\textbf{W}_{\edit{{n}}}\edit{{n}_{j}}\right)
\end{equation}

\begin{equation}
    \label{eq:gat_softmax}
    \alpha_{ij}={\rm softmax}_j(e_{ij})=\frac{{\rm exp}(e_{ij})}{\sum_{k\in \edit{\mathscr{N}}_i}{\rm exp}(e_{ik})}
  \end{equation}
where $e_{ij}$ are the attention coefficients indicating the importance of the features of the node $n_j$ to the node $n_i$, and $\alpha_{ij}$ is normalized across all structure neighbors $\mathscr{N}_i$ of the node $n_i$.
The attention mechanism is responsible for selectivity with node interdependence, which enables us to show how the nodes take effect during the reasoning.

Considering the 10 different types of edges defined in Section \ref{subsec:edge-def}, we model the relational edges basing on the vanilla GAT~\cite{gat}:

\begin{equation}
    \edit{{n}}_i^{l+1}=\frac{1}{\edit{\mathscr{K}}}\parallel_{\edit{\mathscr{k}}=1}^{\edit{\mathscr{K}}}\sigma\left(\sum_{j\in\edit{\mathscr{N}}_{i}}{\sum_{r\in\edit{\mathscr{R}}_{ij}}{\frac{1}{\lvert\edit{\mathscr{N}}_i^r\rvert}\alpha_{r_{ij}}^{k,l}}\edit{\textbf{W}}_{r_{ij}}^{k,l}\edit{{n}}_j^l}\right)
\end{equation}
where $\edit{{n}}_i^l\in\mathbb{R}^{1\times2d}$ is the hidden state of the node $n_i$ in the $l$-th layer, \edit{all the GAT layers are parameter shared, $\mathscr{k}$ is the $\mathscr{k}$-th head following \cite{DBLP:conf/nips/VaswaniSPUJGKP17, gat},} $\edit{\mathscr{R}}$ is the set of all types of edges \edit{in $\mathscr{E}$}, and $\alpha_{r_{ij}}^{k, l}$ are normalized attention coefficients computed by the $k$-th attention mechanism with relation $r$, which is presented in \cite{gat}.

Message passing is a key component of our model.
To echo the \edit{selectivity} of \emph{grandmother cells}, we use the attention mechanism to select (\ie activate or deactivate) key node pairs in our reasoning graph, and we empirically regard this process as the reading reasoning in the graph.

\subsubsection{Gating Mechanism}

A previous study \cite{gcn} showed that GNNs suffer from the smoothing problem when calculated by stacking many layers, thus, we overcome this issue by applying question-aware \cite{DBLP:conf/ijcai/TangSMXYL20} and general gating mechanisms \cite{DBLP:conf/icml/GilmerSRVD17} to optimize the procedure.

\begin{equation}
    \edit{\mathscr{H}}_q={\rm BiLSTM}(H_q)
\end{equation}

\begin{equation}
    w_{ij}=\sigma\left(\textbf{W}_q^{\edit{\top}}[\edit{{n}}_i^l\parallel\edit{\mathscr{H}}_{q_{j}}]\right)
\end{equation}

\begin{equation}
    \edit{\alpha^{gate}_{ij}=\frac{{\rm exp}(w_{ij})}{\sum^{\edit{m}}_{k=1}{\rm exp}(w_{ik})}}
\end{equation}

\begin{equation}
    \edit{q}_i^l=\sum_{j=1}^{\edit{m}}{\alpha^{\edit{gate}}_{ij}\edit{\mathscr{H}_{q_{j}}}}
\end{equation}

\begin{equation}
    \beta_i^l=\sigma(\edit{\textbf{W}}_s^{\edit{\top}}[\edit{q}_i^l\edit{||}\edit{{n}}_i^l])
\end{equation}

\begin{equation}
    \tilde{n}_i^l=\beta_i^l\edit{\odot\tanh}{\left(\edit{p}_i^j\right)}+\left(1-\beta_i^l\right)\odot \edit{{n}}_i^l
\end{equation}
where $\edit{{H}}_q$ is the query representation given by a dedicated \bilstm encoder to keep consistency with the dimension of node features $\edit{\mathscr{N}}$, \edit{$j$ indicates the order of query words, $m$ is the query length}, \edit{$\sigma$ is a sigmoid function}, and $\odot$ indicates element-wise multiplication. Then the general gating mechanism is introduced as follows:

\begin{equation}
    \edit{x}_i^l=\sigma(\edit{\textbf{MLP}}[\edit{{\tilde{n}}_i^l||{n}_i^l}])
\end{equation}

\begin{equation}
    \edit{{n}_i^{l+1}}=\edit{x}_i^l\odot\tanh{\left(\edit{{\tilde{n}}_i^l}\right)}+\left(1-\edit{x}_i^l\right)\odot \edit{{n}_i^l}
\end{equation}

\subsection{Output Layer}\label{subsec:output-layer}

After updating the node representation, we use two multilayer perceptrons, \edit{$\textbf{MLP}_{can}$ and $\textbf{MLP}_{men}$}, to transform the node features to prediction scores.
All the candidate nodes (\emph{grandmother cells}) \edit{$\mathscr{N}_{can}$ and mention nodes $\mathscr{N}_{men}$} from \edit{$\mathscr{G}$} are employed to output the prediction score distribution $a$ as:

\begin{equation}
\begin{aligned}
        a={\gamma\times \edit{\textbf{MLP}}}_{can}\left(\edit{\mathscr{N}}_{can}\right)+\left(1-\gamma\right)\times\max(\edit{\textbf{MLP}}_{men}(\edit{\mathscr{N}}_{men}))
\end{aligned}
\label{eqa:gamma}
\end{equation}
where $\max(\cdot)$ takes the maximum mention node score over $\edit{\textbf{MLP}_{men}}$, then the two parts are summed with the effect of a harmonic $\gamma$ as the final prediction score distribution.

\section{Experiments}\label{sec:4}

We present the performance of our model on the \qangaroo \cite{qan} dataset and evaluate the performance in detail.
Then, the ablation study and the visualization will demonstrate the benefit of the model.
Finally, a case study shows the relationship between the answers output from the models and human reading results.

\subsection{Dataset for Experiments}

\qangaroo is a multi-hop MRC dataset containing two independent datasets, \WikiHop and \MedHop, from the open-domain field and molecular biology field, respectively.
Both \WikiHop and \MedHop were divided into three subsets: the training set, development set, and undisclosed test set, which is used for official evaluation.
The dataset sizes are shown in Table~\ref{tab:dataset-size}.

\begin{table}[ht]
    \caption{Dataset Size of \WikiHop and \MedHop}
    \centering
    \begin{tabular}{ccccc}
    \toprule
            & Training  & Development & Test  & Total  \\
    \midrule
    WikiHop & 43,738 & 5,129       & 2,451 & 51,318 \\
    MedHop  & 1,620  & 342         & 546   & 2,508  \\
    \bottomrule
    \end{tabular}
    \label{tab:dataset-size}
\end{table}

\begin{figure*}[ht]
        \centering
        \subfigure[A sample from the \WikiHop.]{
        \begin{minipage}[t]{0.4\linewidth}
        \centering
        \includegraphics[width=\linewidth]{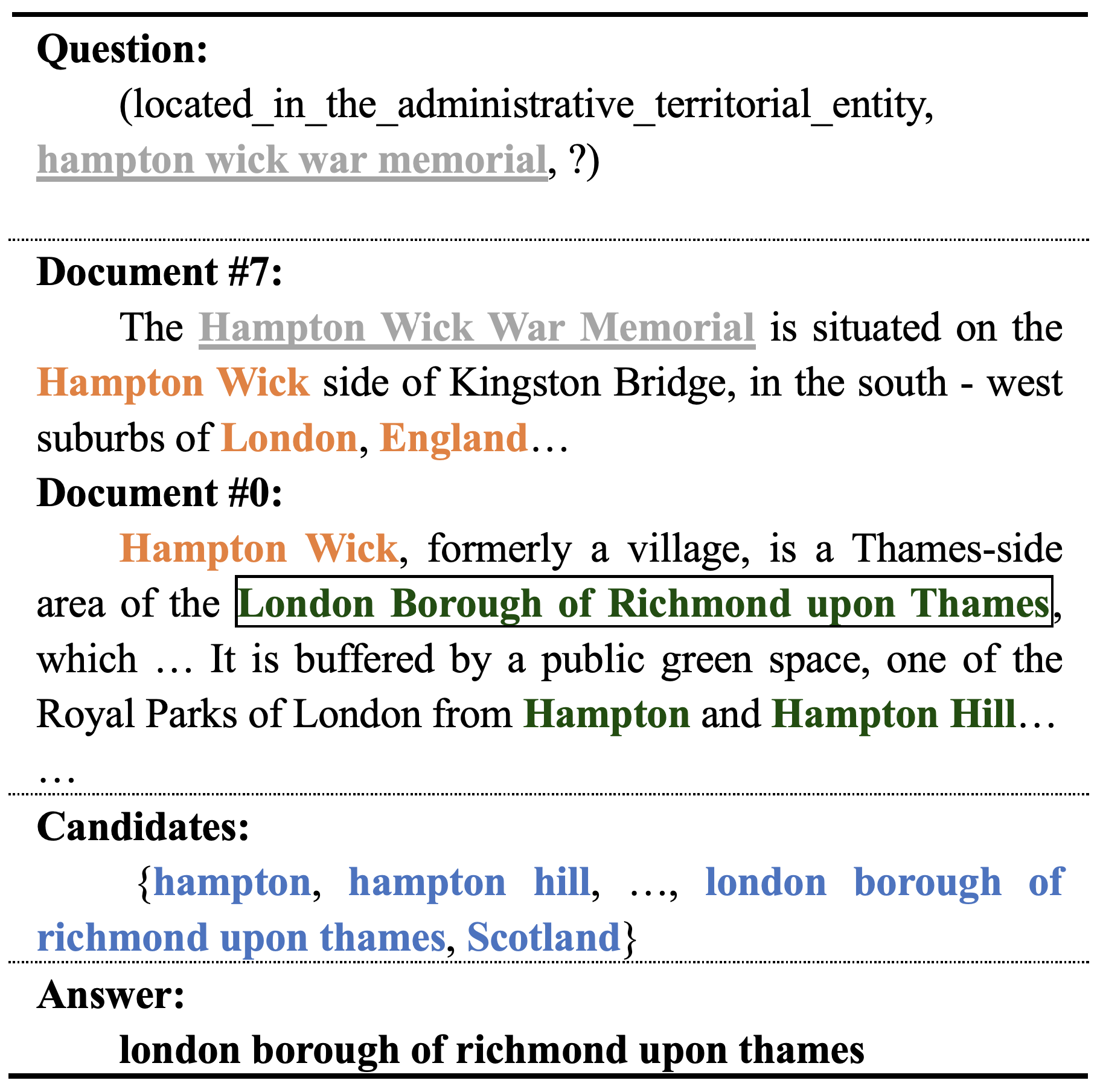}
        \label{fig:sample-wikihop}
        \end{minipage}%
        }%
        \hspace{1cm}
        \subfigure[A sample from the \MedHop.]{
        \begin{minipage}[t]{0.4\linewidth}
        \centering
        \includegraphics[width=\linewidth]{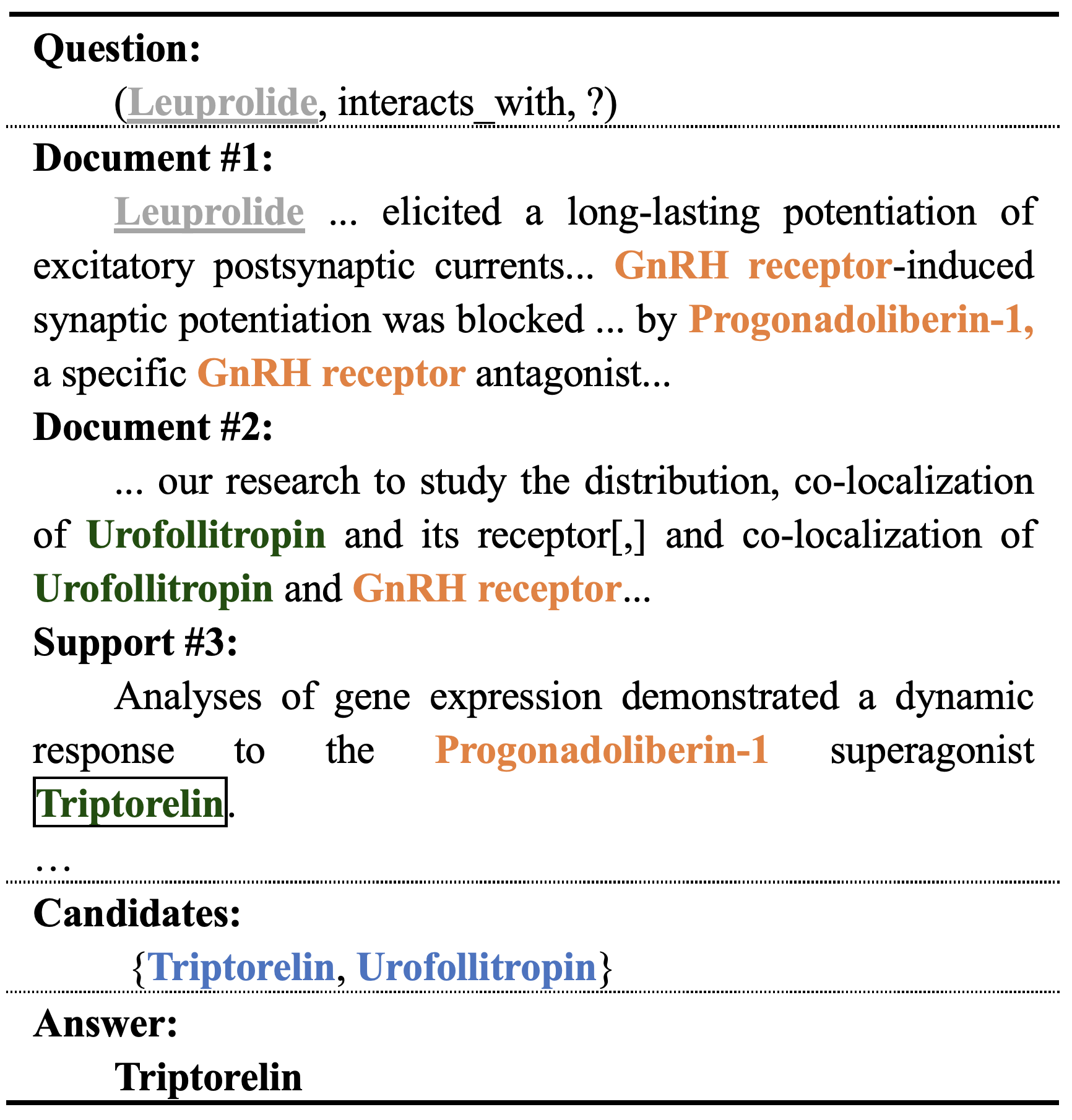}
        \label{fig:sample-medhop}
        \end{minipage}%
        }%
        \centering
        \caption{Samples of \WikiHop and \MedHop. Subject entities, reasoning entities, mention entities, and candidate entities are shown in gray, orange, green, and blue colors, respectively. The occurrence of the correct answer is shown by a square frame outside.}
        \end{figure*}

\WikiHop was created from \Wikipedia (as the document corpus) and \Wikidata (as structured knowledge triples).
A sample from the dataset is shown in Figure~\ref{fig:sample-wikihop}.
In this sample, the query (\emph{located\_in\_the\_administrative\_territorial\_entity, hampton\_wick\_war\_memorial, ?}) requires us to answer the administrative territory of the \emph{Hampton Wick War Memorial}.
To predict it, a named recognition entity \emph{Hampton Wick} is extracted from the seventh support document, and it links to the same tokens in the zeroth support document where the correct candidate answer appears as well.
The reasonable clue chain \emph{Hampton Wick War Memorial $\leftrightarrow$ Hampton Wick \# 1 $\leftrightarrow$ Hampton Wick \# 2 $\leftrightarrow$ London Borough of Richmond upon Thames} presents the procedure of our model for the multi-hop MRC task.

To validate whether the dataset can be consistent with the formalization of the multi-hop MRC, the dataset founder asked human annotators to evaluate the samples in the \WikiHop development and test sets.
For each sample in the two sets, at least three annotators participated in the evaluation, and they were required to answer three questions:
\begin{itemize}
    \item whether they knew the fact before
    \item whether the fact follows from the texts (with options \emph{follows}, \emph{likely}, and \emph{not follows})
    \item whether multiple documents are required to answer the question
\end{itemize}

All the samples in the test set were human-selected and were labeled by the majority of annotators with \emph{follows} and \emph{multiple documents required}.
Annotators merely noted the samples in the development set without the selection.

The \MedHop dataset was constructed using the \DrugBank as certain knowledge. Then the creators extracted the research paper abstracts from \MEDLINE --- the online medical literature search \& analysis system and the bibliographic database of the National Library of Medicine of the USA --- as a corpus, and the aim is to predict the drug-drug interaction (DDI) after reading the texts.
The purpose of applying multi-hop methods in this prediction is to find and combine individual observations that can suggest previously unobserved DDI from inferring and reasoning the prior public knowledge in contents rather than some costly experiments.
The only query type is \emph{interacts\_with}.
A sample given in \cite{qan} is illustrated in Figure~\ref{fig:sample-medhop} and note that accession numbers replace the medical proper nouns (e.g., DB00007, DB06825, DB00316) rather than the names of drugs and human proteins (e.g., \emph{Leuprolide}, \emph{Triptorelin}, \emph{Acetaminophen}) in practice.

\subsection{Experiments Settings}

We exploited \nltk \cite{DBLP:conf/acl/Bird06} toolkit to tokenize the support documents and candidates, then split the query $q=\{s, r, a^*\}$ into relation $r$ and subject entity $s$.
All the named entities matching with candidates $C_q$ were extracted as mention nodes $\edit{\mathscr{V}}_{men}$, and the \spacy \footnote{https://spacy.io} was used to extract the named entities and noun phrases from texts as reasoning nodes $\edit{\mathscr{V}}_{rea}$.
We concatenated \glove \cite{pennington2014glove} and n-gram character embeddings \cite{DBLP:conf/emnlp/HashimotoXTS17} to obtain 400-dimensional word embeddings, which were input to the encoder layer. The out-of-vocabulary words were presented with random vectors.
The word embedding was fixed in \WikiHop experiment and trainable on \MedHop.
We implemented the \crname model with \pytorch and \pyg \cite{DBLP:journals/corr/abs-1903-02428}. \nx \cite{Schult08exploringnetwork} was utilized to visualize the reading graph, the weights of node pair weights, and node selections.

\subsection{Results and Analyses}

In Table~\ref{tab:modelresult} we present the performance of \crname in the development and test sets of \WikiHop and \MedHop and compare it with the performance of published models mainly based on GNNs.
Our model improved the accuracy of GCN-based models \HDE \cite{hde} in the test set from 70.9\% to 72.0\% and \PathbasedGCN in the development set from 64.5\% to 66.9\%, while \PathbasedGCN using \glove and \ELMo word embeddings surpassed our model by 0.5\% in the test set, which confirms that the initial representations of nodes are extremely critical \cite{DBLP:conf/ijcai/TangSMXYL20}.
However, limited by the architecture and computing resources, we did not use powerful contextual word embeddings like \ELMo and \bert in our model, which can be further addressed.
Compared to the other GNN-based models \cite{DBLP:conf/naacl/CaoAT19, DBLP:conf/naacl/CaoFT19, 10.1109/TKDE.2020.2982894} and the sequential models \cite{qan, DBLP:conf/naacl/DhingraJYCS18}, our model achieved higher accuracy.
We are the first to apply the GNN-based model to \MedHop, although the accuracy was 1.8\% lower than \BiDAF, we believe that the possible reason was the failure in extracting the reasoning nodes of the \spacy toolkit, which means the bridge entities were incomplete.

\begin{figure}[]
    \centering
    \includegraphics[width=\linewidth]{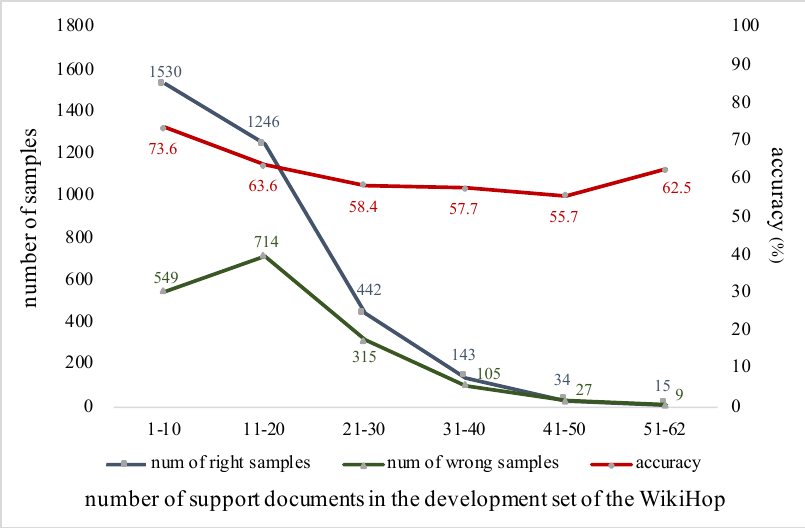}
    \caption{
        Statistics of the model performance with different numbers of support documents on the \WikiHop development set.
    }
    \label{fig:numaccwiki}
    \end{figure}

\begin{figure}[]
    \centering
    \includegraphics[width=\linewidth]{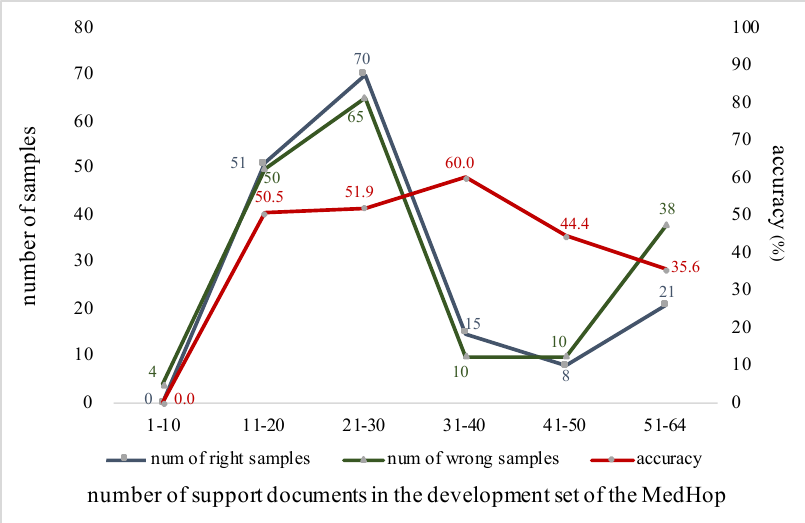}
    \caption{
        Statistics of the model performance with different number of support documents on the \MedHop development set.
    }
    \label{fig:numaccmed}
    \end{figure}

To analyze the scalability of our model, we divided the development set into six groups according to the number of support documents and then determined the accuracy in each group.
The grouped accuracy on \WikiHop is shown in Figure~\ref{fig:numaccwiki}. \crname achieved competitive results: 73.59\% and 63.57\% in the groups of (1-10) and (11-20), with a total of 4,039 samples accounting for 95\% of the development set.
The lowest accuracy of 55.74\% was for the group (41-50). However, it increased to 62.5\% in the group (51-62), which shows the scalability of our model is effective.
The grouped accuracy on \MedHop is shown in Figure~\ref{fig:numaccmed}, and they are quite competitive. The highest and second-highest accuracy of 60.00\% and 51.85\% are in (31-40) and (21-30) groups, respectively, and the lowest and second-lowest accuracy of 0\% and 35.59\% are in (1-10) and (51-62) groups, respectively.
In particular, the result in the (51-64) group on \MedHop is against the group (51-62) on \WikiHop, which implies that we must concentrate on the difference between the open-domain and molecular textual contexts.
The results in the different number of support documents show the contribution of our model to the scalability of the multi-hop MRC tasks.

\begin{table*}[]
    \caption{Performance of the proposed \crname in the development and test sets of \WikiHop and \MedHop, and comparisons with other published approaches on the leaderboard.}
    \centering
    \begin{tabular}{lcccc}
    \toprule
    \multirow{2}{*}{Single models}  & \multicolumn{2}{c}{WikiHop Accuracy (\%)} & \multicolumn{2}{c}{MedHop Accuracy (\%)} \\
                                    & Dev                 & Test                & Dev                 & Test               \\ \midrule
    Coref-GRU \cite{DBLP:conf/naacl/DhingraJYCS18}               & 56.0                & 59.3                & -                   & -                  \\
    MHQA-GRN \cite{10.1109/TKDE.2020.2982894}                & 62.8                & 65.4                & -                   & -                  \\
    Entity-GCN \cite{DBLP:conf/naacl/CaoAT19}              & 64.8                & 67.6                & -                   & -                  \\
    HDE \cite{hde}                    & 68.1       & 70.9                & -                   & -                  \\
    BAG \cite{DBLP:conf/naacl/CaoFT19}                    & 66.5                & 69.0                & -                   & -                  \\
    Path-based GCN \cite{DBLP:conf/ijcai/TangSMXYL20} & 64.5                & -                   & -                   & -                  \\
    Document-cue \cite{qan}            & -                   & 36.7                & -                   & 44.9               \\
    FastQA \cite{qan}                  & -                   & 25.7                & -                   & 23.1               \\
    TF-IDF \cite{qan}                  & -                   & 25.6                & -                   & 9.0                \\
    BiDAF \cite{qan}                  & -                   & 42.9                & -                   & 47.8      \\
    \crname                      & 66.5                & 72.0      & 48.2       & 46.0               \\ \bottomrule
    \end{tabular}
    \label{tab:modelresult}
\end{table*}

As mentioned above, the \WikiHop development set had consistency between facts and annotated documents. To determine whether multiple documents are required to reason the question, we split our models into five categories as follows. In each category, all three annotators annotated:
(1) \emph{requires multiple documents} and \emph{follows fact};
(2) \emph{requires single document} and \emph{follows fact};
(3) \emph{requires multiple documents} and \emph{likely follows fact};
(4) \emph{requires single document} and \emph{likely follows fact};
(5) \emph{not follows} is not given.
The performance of our model is presented in Table~\ref{tab:wikihop-sub-ana}.
We observe that \crname had the best performance of 74.9\% in the samples which follow the facts and require multiple passages.
This phenomenon proves the effectiveness of the model in pure multi-hop MRC tasks.
It achieved the second-best result of 74.0\% in samples following the facts and requiring a single document, which supports that \crname is also effective in single-passage MRC tasks.
Further, we believe that authenticity can seriously impact the accuracy of our prediction. The categories associated with \emph{may not follow the fact} achieved the worse results, of 71.4\%, 71.4\%, and 71.5\%, respectively, in the groups of \emph{likely follows the fact (single document} and \emph{multiple documents)} and \emph{``not follows'' is not given}.
The same analysis is infeasible in the development set of \MedHop since the document complexity and the number of documents per sample are significantly larger.

\begin{table}[]
    \caption{Performance on the \WikiHop development subset}
    \centering
    \begin{tabular}{l|l|c}
    \toprule
    \multicolumn{2}{c|}{Annotation}                                   & \begin{tabular}[c]{@{}c@{}}Accuracy (\%)\end{tabular} \\ \midrule
    \multirow{2}{*}{follows fact}        & requires\;multiple\;documents & \textbf{74.9}                                            \\
                                         & requires\;single\;document   & 74.0                                                     \\ \hline
    \multirow{2}{*}{likely follows fact} & requires\;multiple\;documents & 71.4                                                     \\
                                         & requires\;single\;document   & 71.4                                                     \\ \hline
    \multicolumn{2}{c}{\emph{not follows} is not given}                     & 71.5                                                     \\ \bottomrule
    \end{tabular}
    \label{tab:wikihop-sub-ana}
    \end{table}

\subsection{Ablation Study}

We proposed five types of nodes in $\edit{\mathscr{G}}$, to analyze how they reasoned, we removed the edges with specific connections and isolated the nodes to evaluate the performance in the subset of the \WikiHop development set, that is, \emph{not follows} was not annotated.
Moreover, we tested the model without the message passing in $\edit{\mathscr{G}}$. The ablated performance is shown in Table~\ref{tab:ab-study}.

\begin{table}[ht]
\caption{Ablation Performance on the \qangaroo Development Subset}
    \centering
\begin{tabular}{lcccc}
\toprule
\multicolumn{1}{l}{\multirow{2}{*}{Model}} & \multicolumn{4}{c}{Accuracy (\%)}                  \\ \cline{2-5}
\multicolumn{1}{c}{}                       & \WikiHop & $\Delta$             & \MedHop & $\Delta$             \\ \midrule
\multicolumn{1}{l}{Full Model}             & 71.45   & -              & 48.25  & -              \\
w/o\;GAT                                    & 52.69   & \textbf{18.76} & 37.72  & 10.53          \\
w/o\;$N_{sub}$                                     & 70.95   & 0.5            & 47.37  & 0.88           \\
w/o\;$N_{men}$                                     & 63.34   & 8.11           & 4.97   & \textbf{43.28} \\
w/o\;$N_{rea}$                                     & 70.77   & 0.68           & 47.37  & 0.88           \\
w/o\;$N_{sup}$                                     & 62.02   & 9.43           & 48.54  & -0.29          \\
w/o\;$N_{can}$                                     & 65.87   & 5.58           & 44.77  & 3.48           \\ \bottomrule
\end{tabular}
\label{tab:ab-study}
\end{table}

\begin{table}[ht]
    \caption{Ablation studies of hyperparameters of \edit{GAT layers} and \edit{weights of \emph{grandmother cells}} in reasoning graph predictions.}
    \centering
    \begin{tabular}{cccc}
    \toprule
    Hyperparameters                                 & value & Acc. of \WikiHop                    & Acc. of \MedHop \\ \midrule
    \multirow{4}{*}{\edit{$l$}}                 & 3     & \multicolumn{1}{c}{57.8}          & 42.4             \\
                                                    & 4     & \multicolumn{1}{c}{58.5}          & 43.3             \\
                                                    & 5     & \multicolumn{1}{c}{\textbf{66.5}} & \textbf{48.2}             \\
                                                    & 6     & \multicolumn{1}{c}{64.2}          & 45.0             \\ \midrule
    \multirow{4}{*}{$\gamma$} & 0     & \multicolumn{1}{c}{59.7}          & 42.7             \\
                                                    & 0.5   & \multicolumn{1}{c}{66.1}          & 44.2             \\
                                                    & 1.0   & \multicolumn{1}{c}{\textbf{66.5}} & \textbf{48.2}             \\
                                                    & 1.5   & \multicolumn{1}{c}{59.1}          & 43.3             \\ \bottomrule
    \end{tabular}
    \label{numofhopandalpha}
\end{table}

On \WikiHop, the proposed heterogeneous graph attention network was the most effective component of \crname. Without its contribution, the accuracy decreased by 18.76\%.
After blocking the nodes by groups, we observed that the support nodes contributed 9.43\% absolutely, the mention nodes dedicated 8.11\% and the candidate nodes contributed 5.58\%.
Regarding the reasoning and subject nodes, we considered the small quantities contained in the graph leading to low status in contributions.
However, we observed considerably different performances between \WikiHop and \MedHop.
As the results are shown in Table~\ref{tab:ab-study}, the most effective part of the model is mention nodes. When we blocked the mention nodes in the graph, the accuracy decreased significantly, by 43.28\%, and the graph reasoning contributed 10.53\% to accuracy.
Meanwhile, support nodes had negative effects on the prediction, a decrease of 0.29\%, which is diametrically opposite the performance on the \WikiHop development subset.

In Table~\ref{numofhopandalpha}, we present the model performances with different hyperparameters, especially the number of stacked GAT layers (the number of hops) and the weight of \emph{grandmother cells}.
The number of GAT layers controls how many parameter-sharing GAT layers should be involved in the reasoning graph.
On \WikiHop, we obtained the highest accuracy (66.5\%) when we stacked the graph with five layers, and the model with three or four GAT layers had poorer performance (57.8\% or 58.5\% respectively).
With six GAT layers, the accuracy dropped 2.3\% compared with the best performance.

Furthermore, as the final prediction illustrated in Equation (\ref{eqa:gamma}), $\gamma$ coordinates the mention nodes and the candidate nodes \emph{grandmother cells}; we present the model performances with different $\gamma$ settings in Table~\ref{numofhopandalpha}.
The best performance was with $\gamma$ set to $1$. However, if we gave it too much weight, that is $\gamma=1.5$, the accuracy decreased by 7.4\%, which is even worse than when we set $\gamma$ to $0$ (59.7\%), which convinces us that we should not ignore the effect of much larger networks behind \emph{grandmother cells}.
We observed similar phenomena with different hyperparameter settings On \MedHop.
When the number of hops was 5, and $\gamma$ is $1$, the model performed best at approximately 48.2\%.
We suspect that when a few GAT layers are stacked, the messages of nodes cannot pass sufficiently among the reasoning graph. When too many GAT layers are stacked, the graph over-smoothing problem leads to a drop in accuracy.
We also empirically observed that models with higher $\gamma$ may lose semantic information from context resulting in reduced prediction accuracy, which also fits the concept of \emph{grandmother cells} that before the final predicting determination, a huge background network calculation should be done implicitly.

\begin{figure}[]
    \centering
    \subfigure[]{\includegraphics[width=0.492\linewidth]{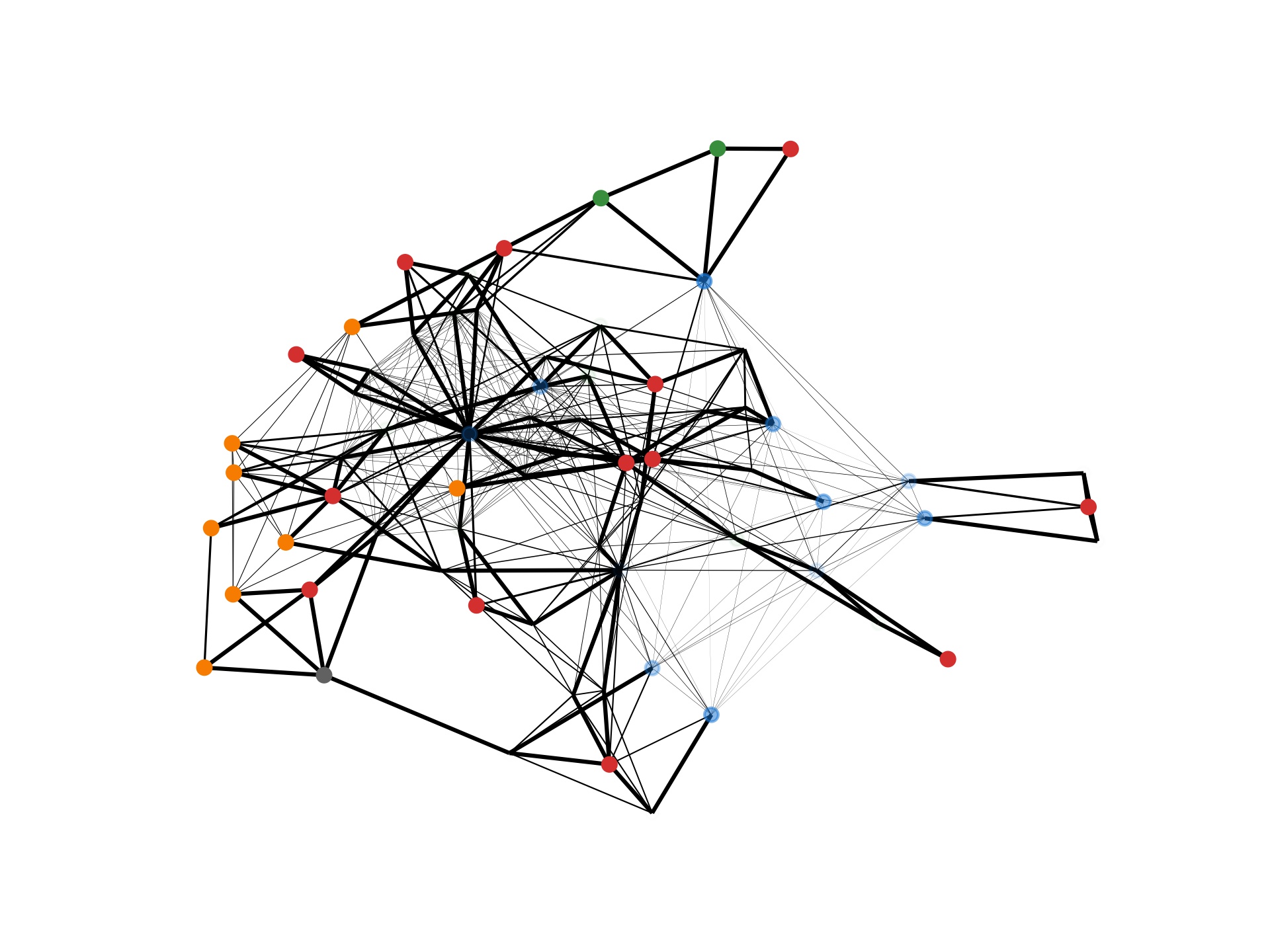}}
    \hfill
    \subfigure[]{\includegraphics[width=0.492\linewidth]{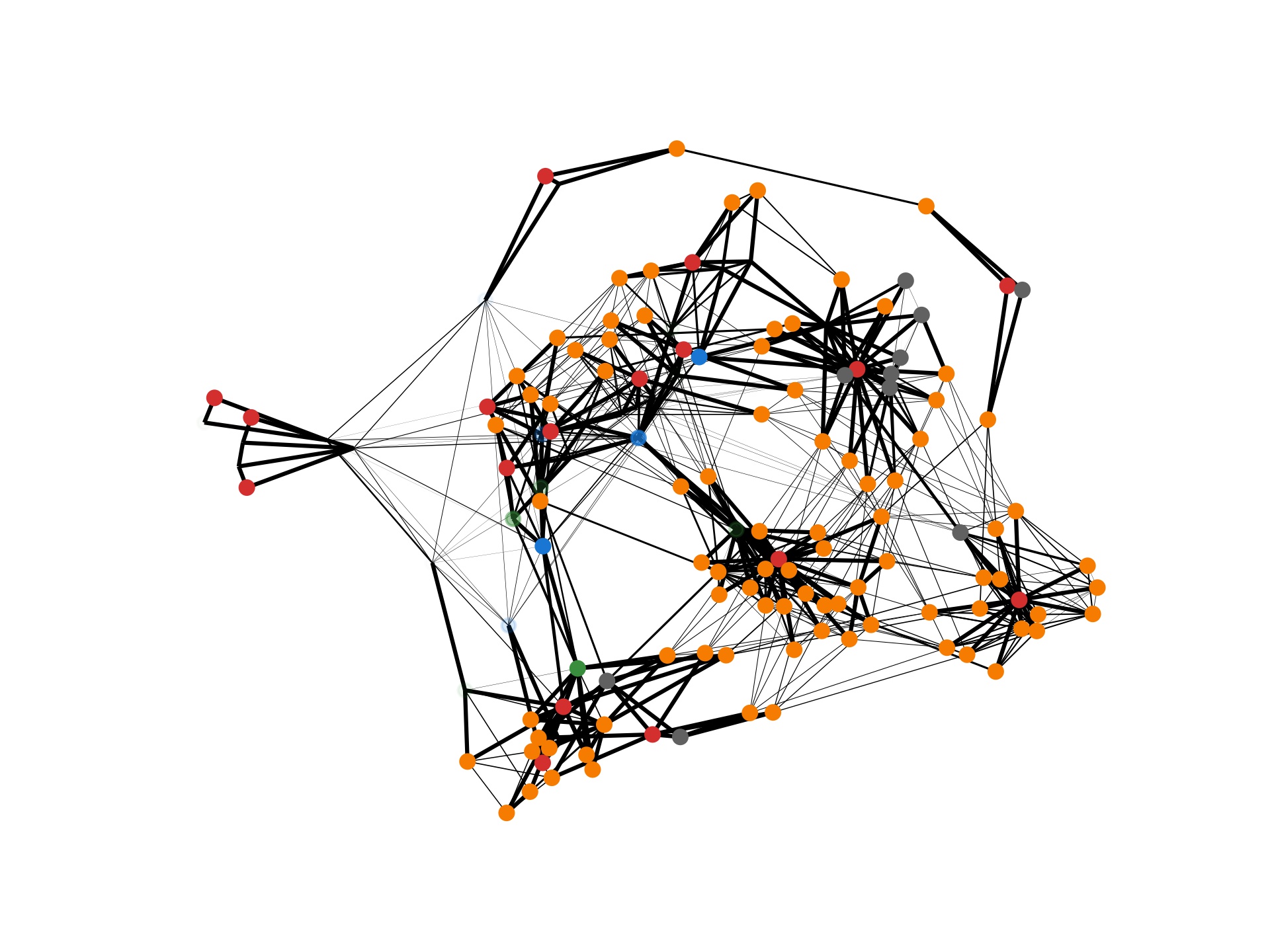}}
    \vfill
    \subfigure[]{\includegraphics[width=0.492\linewidth]{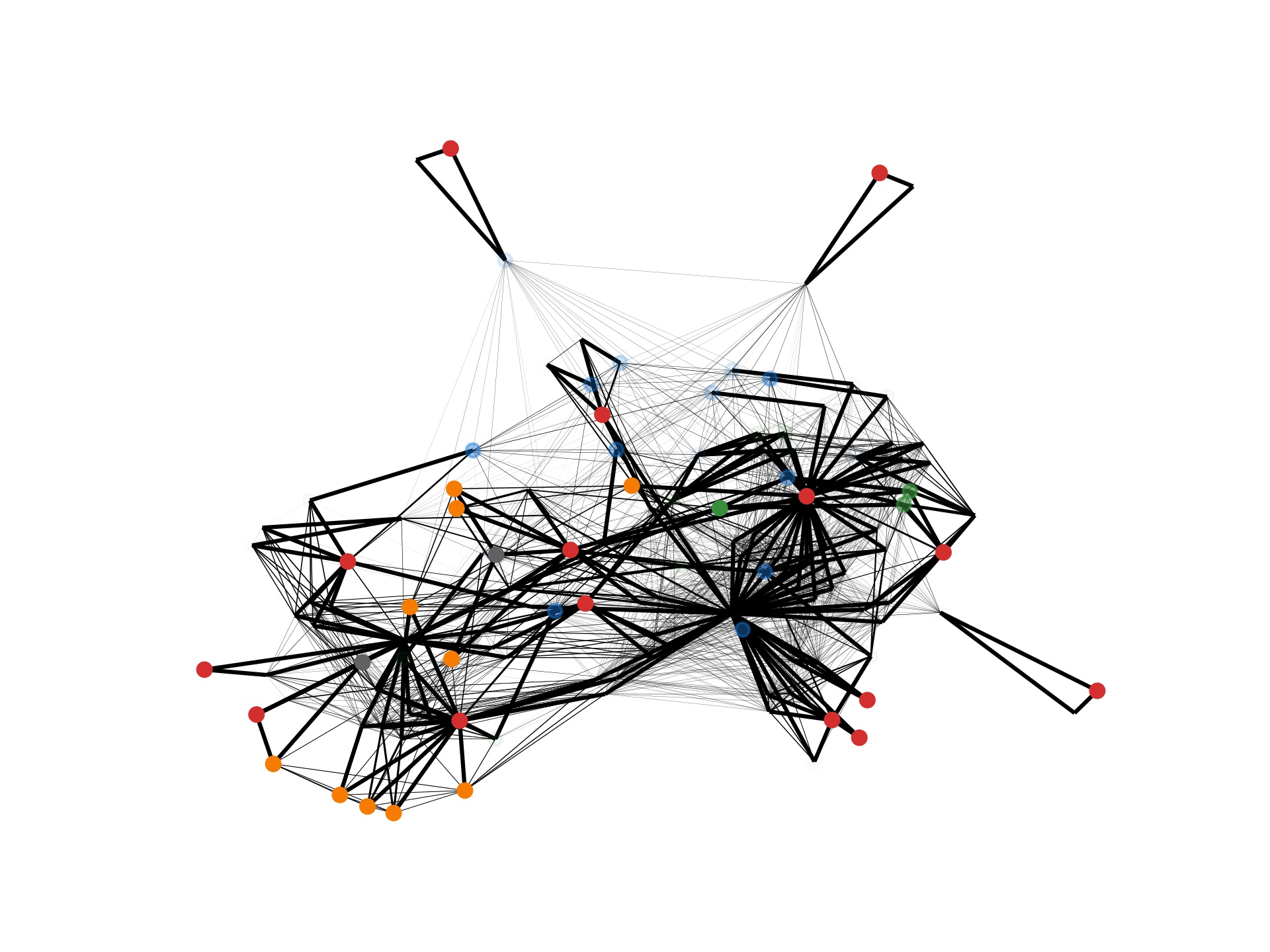}}
    \hfill
    \subfigure[]{\includegraphics[width=0.492\linewidth]{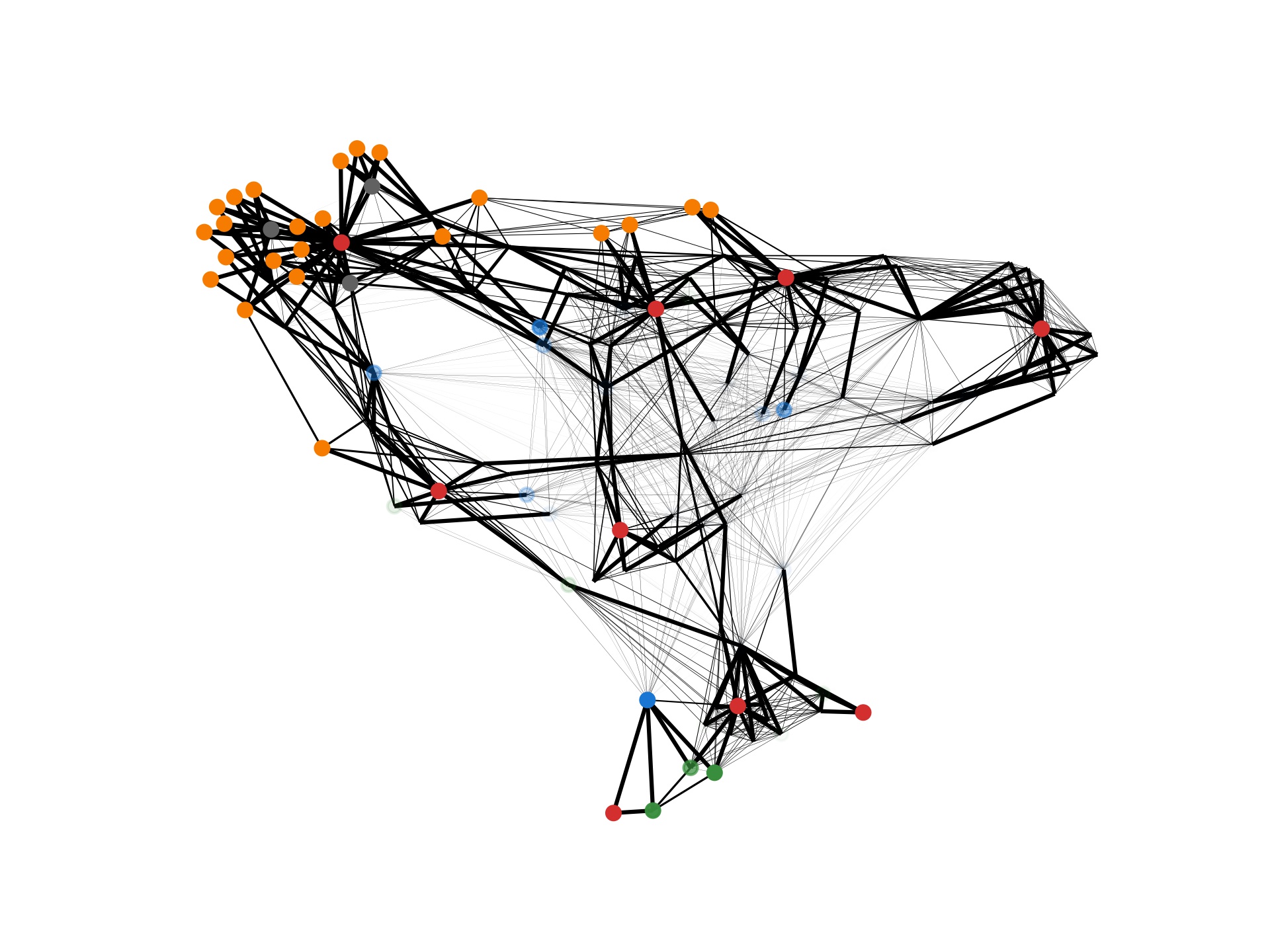}}
    \vfill
    \subfigure[]{\includegraphics[width=0.492\linewidth]{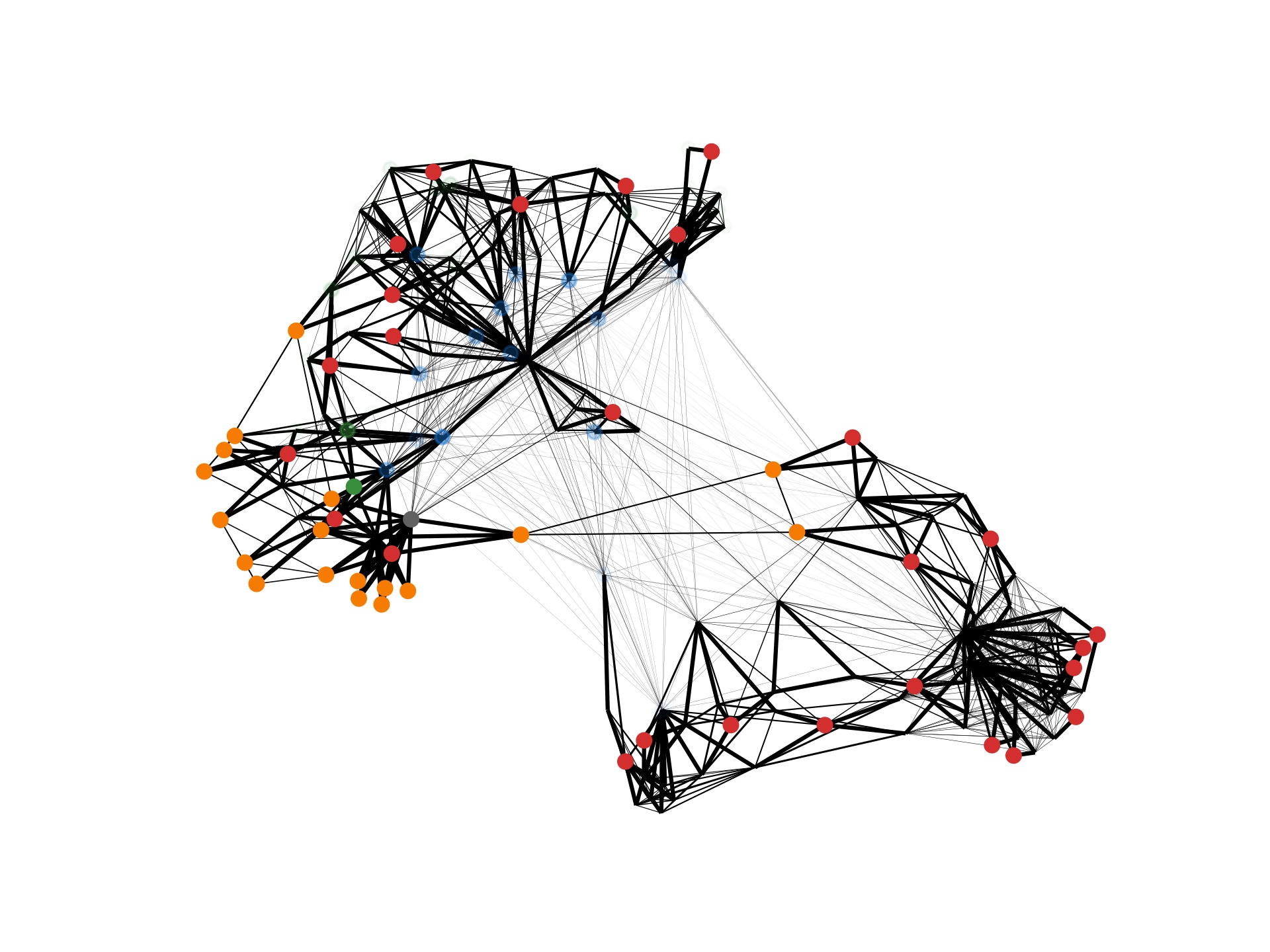}}
    \hfill
    \subfigure[]{\includegraphics[width=0.492\linewidth]{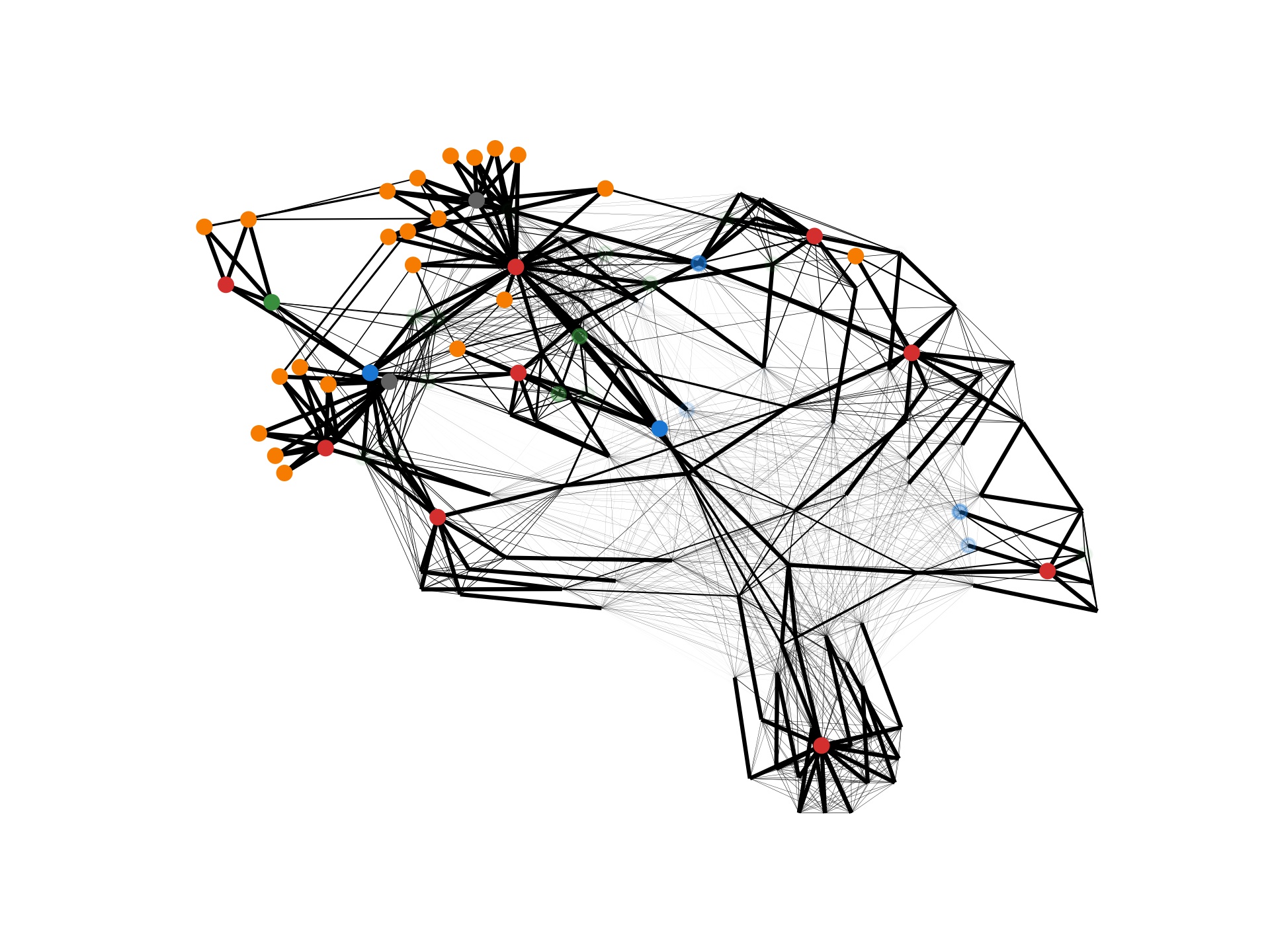}}
    \caption{Visualizations of reasoning graphs on the \WikiHop development samples that are correctly answered. A thicker edge corresponds to a higher attention weight, and darker green nodes or darker blue nodes represent higher output values among the same type of nodes.}
    \label{fig:wikivis}
    \end{figure}

    \begin{figure}[]
        \centering
        \subfigure[]{\includegraphics[width=0.492\linewidth]{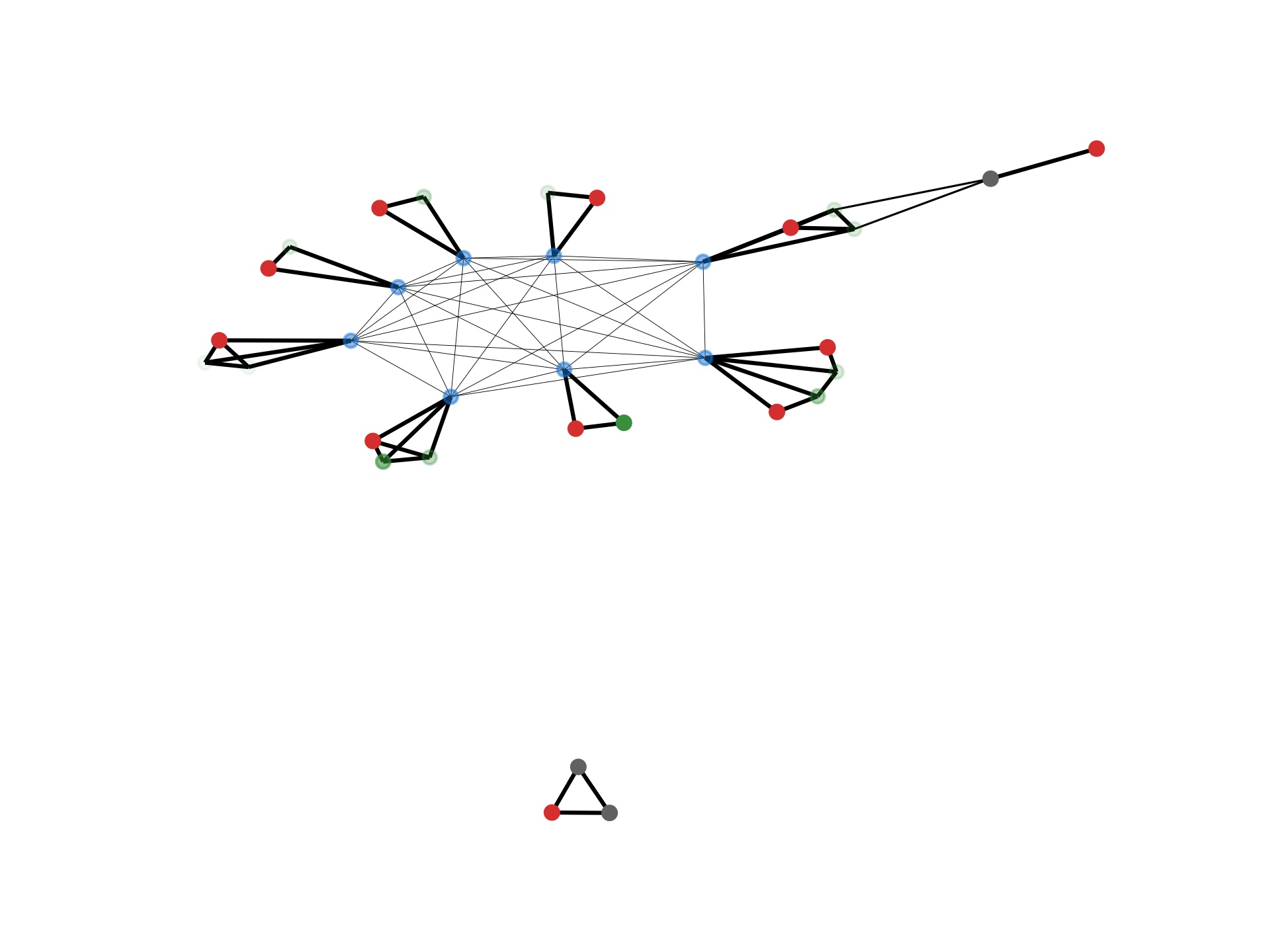}}
        \hfill
        \subfigure[]{\includegraphics[width=0.492\linewidth]{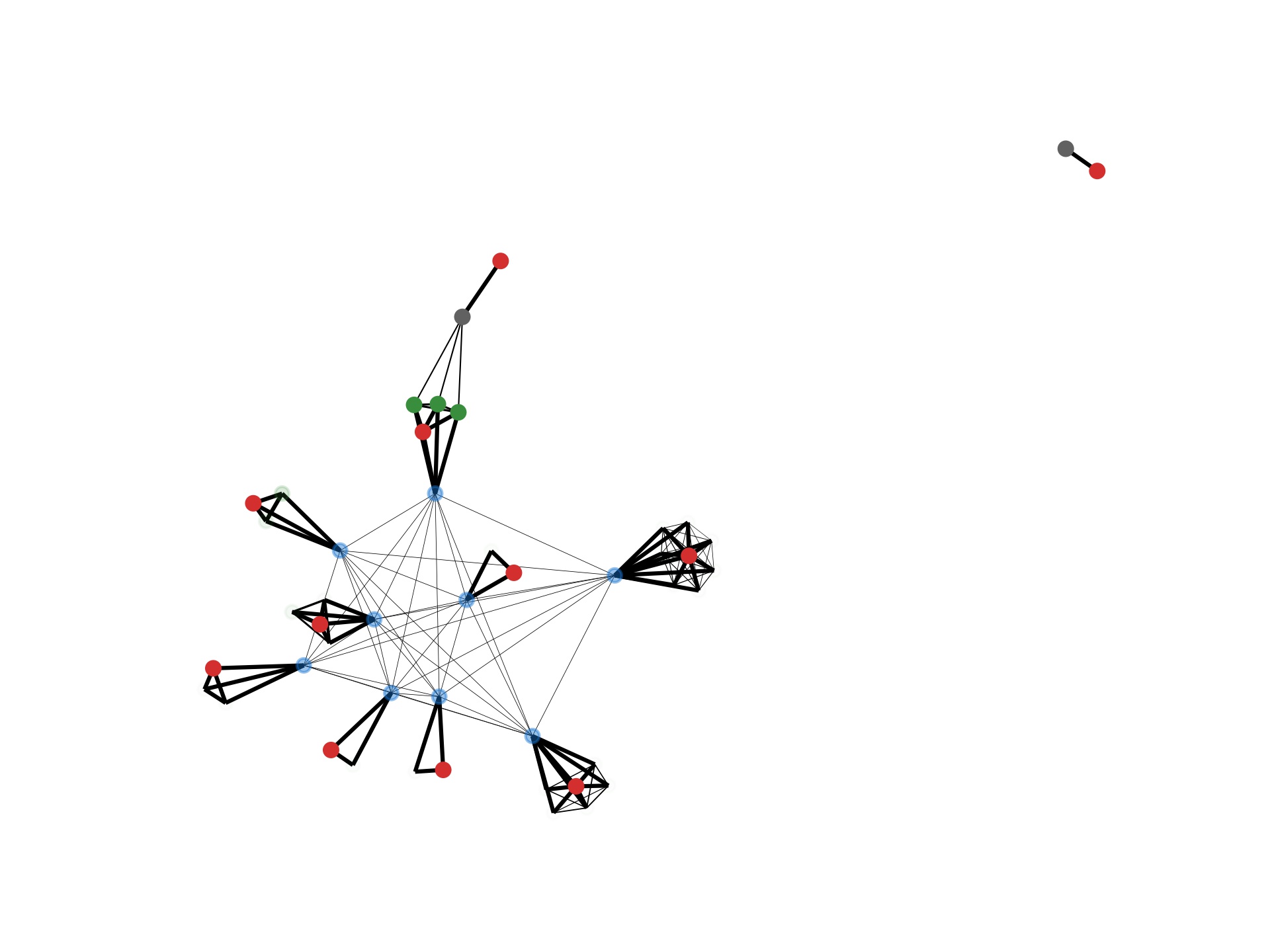}}
        \vfill
        \subfigure[]{\includegraphics[width=0.492\linewidth]{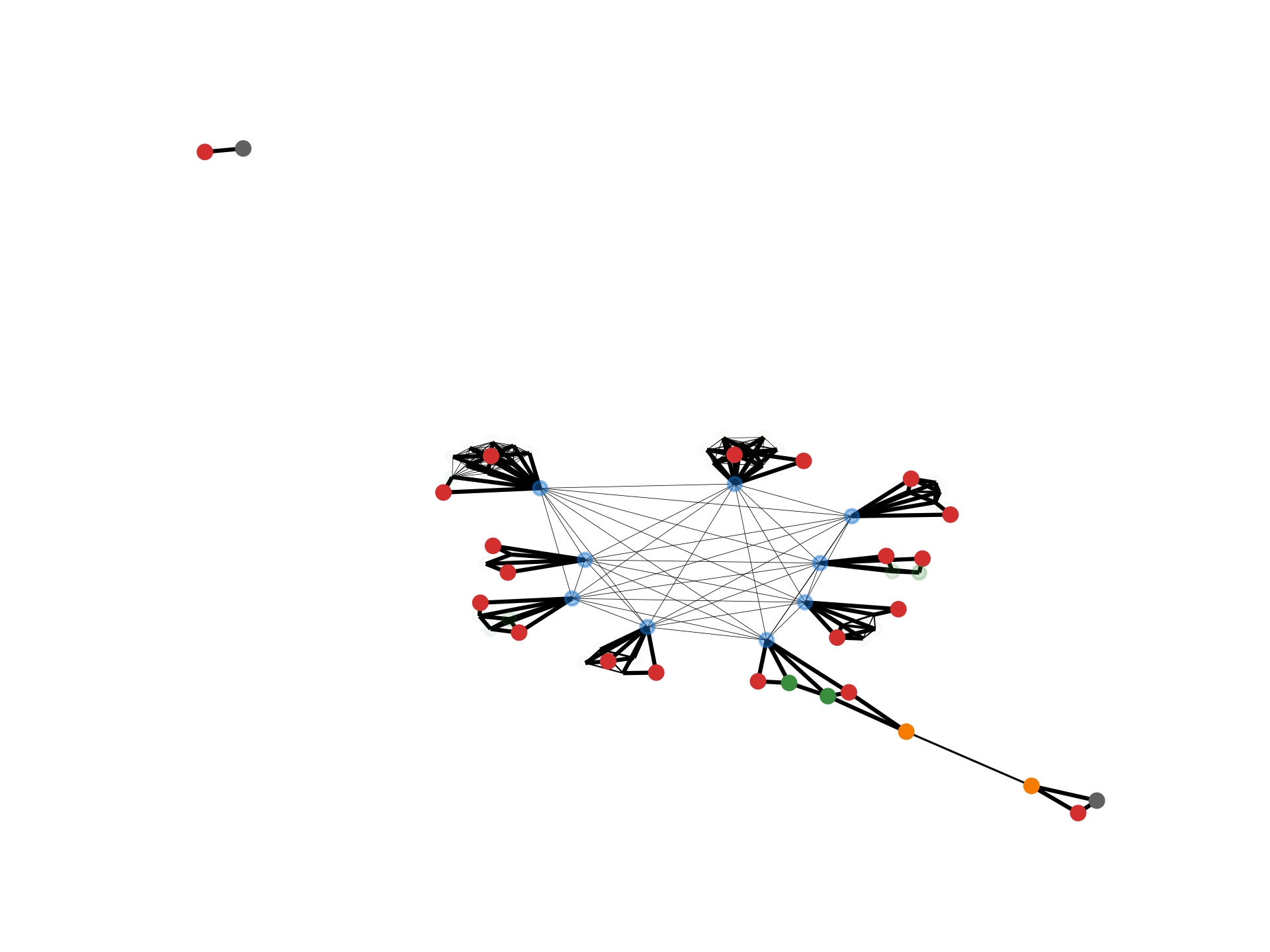}}
        \hfill
        \subfigure[]{\includegraphics[width=0.492\linewidth]{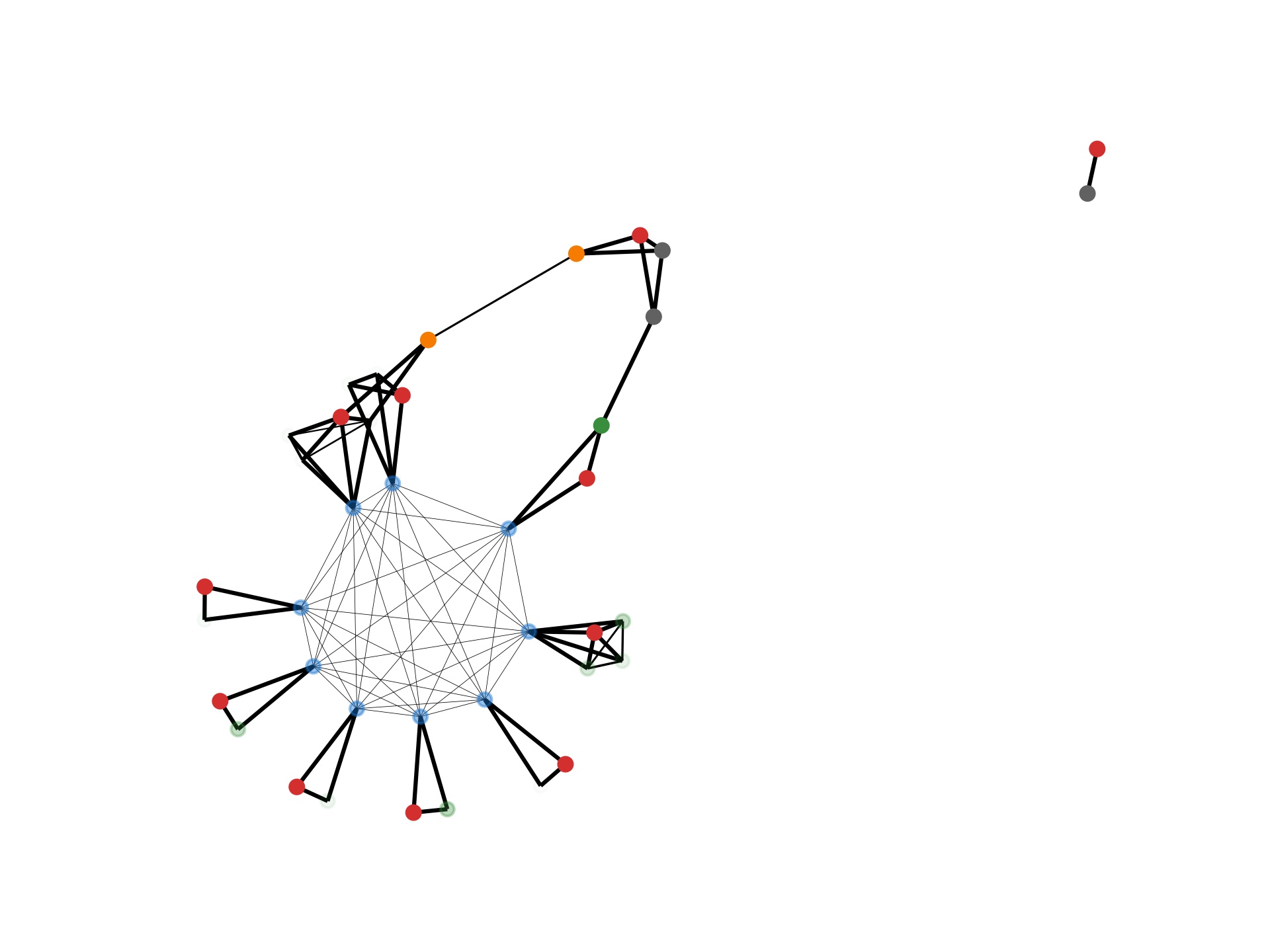}}
        \vfill
        \subfigure[]{\includegraphics[width=0.492\linewidth]{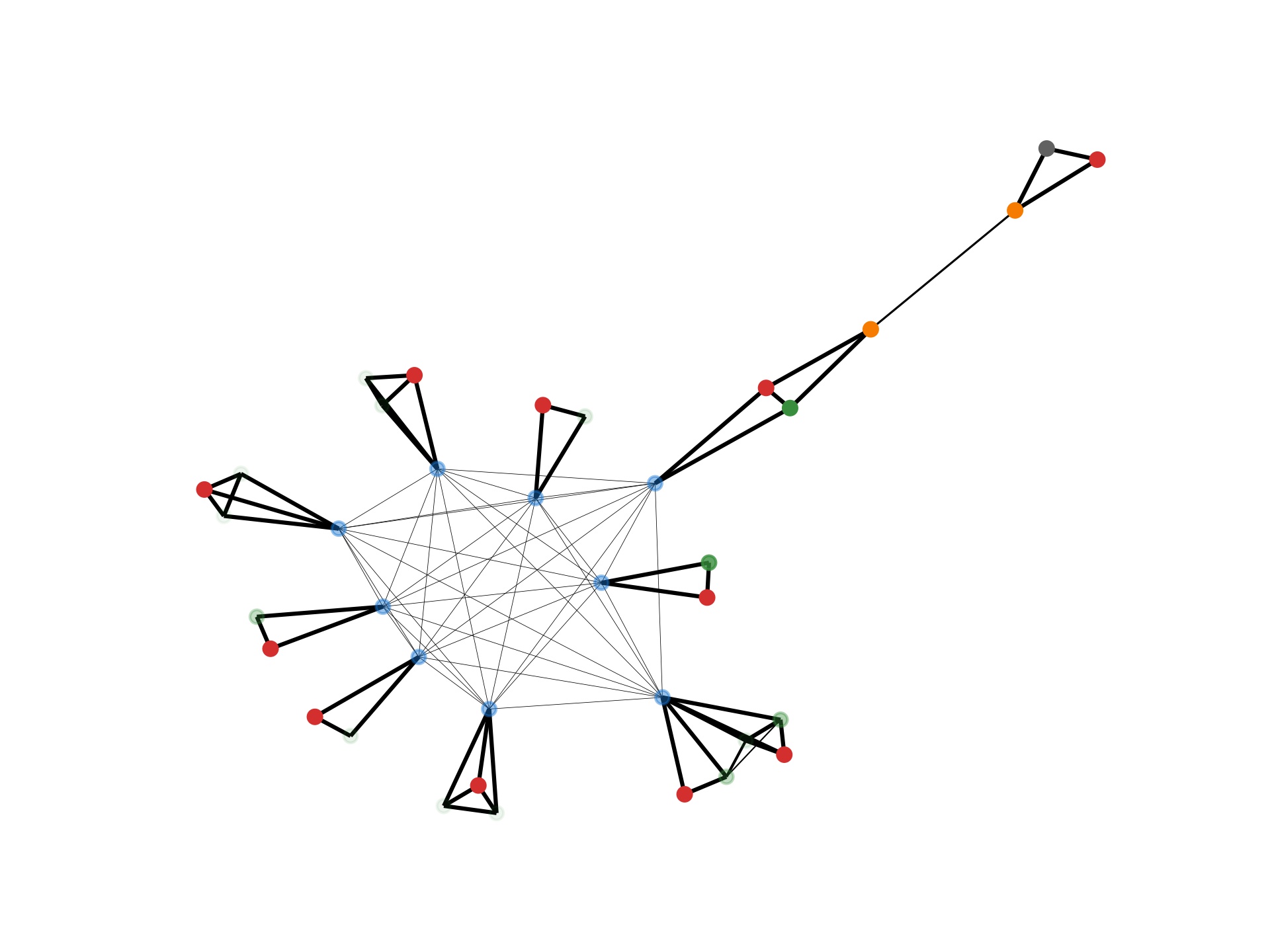}}
        \hfill
        \subfigure[]{\includegraphics[width=0.492\linewidth]{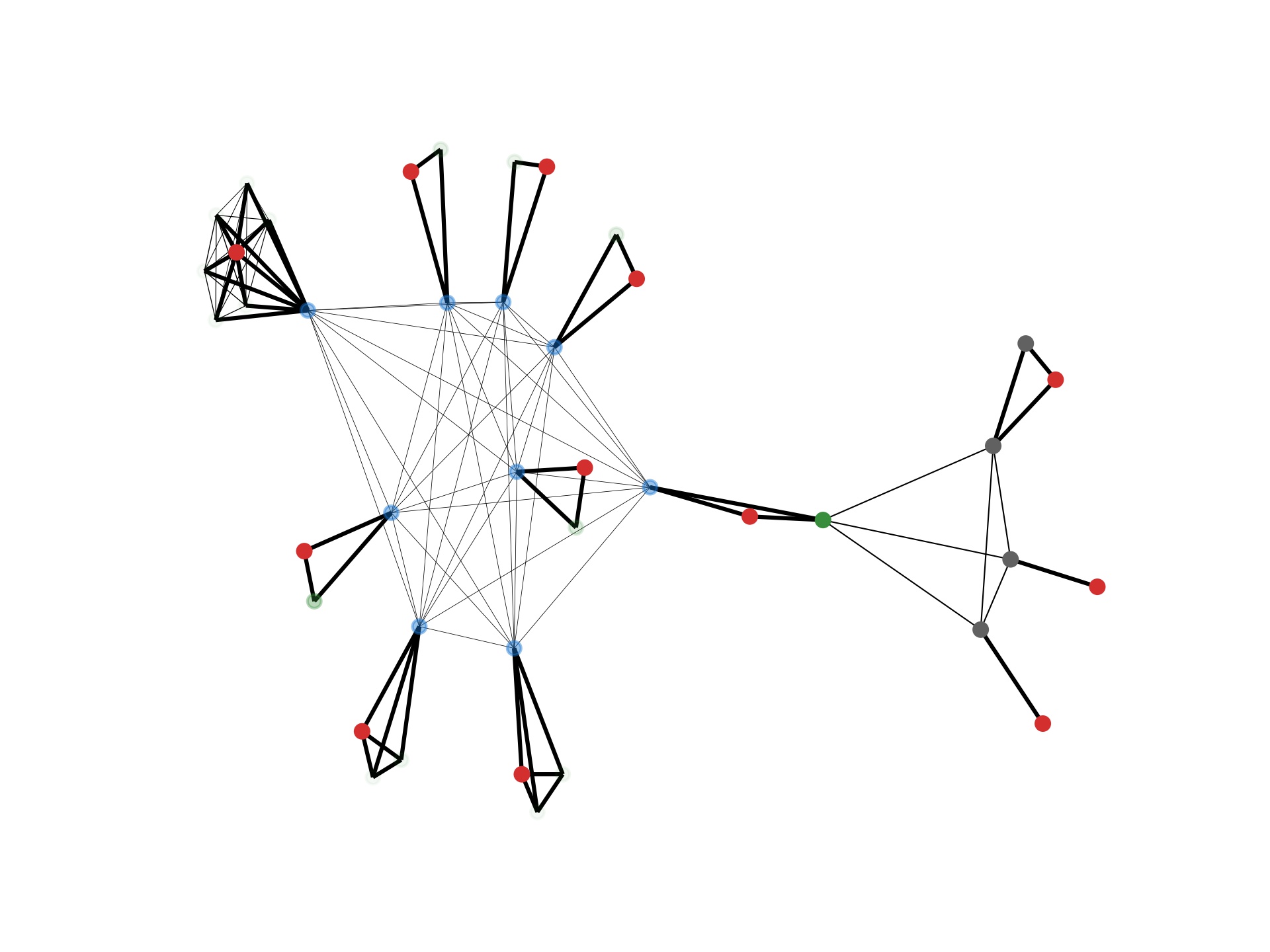}}

        \caption{Visualizations of reasoning graphs on the \MedHop development samples that are correctly answered. A thicker edge corresponds to a higher attention weight, and darker green nodes or darker blue nodes represent higher output values among the same type of nodes.}
        \label{fig:medvis}
        \end{figure}

\begin{figure}[ht]
    \centering
    \includegraphics[width=\linewidth]{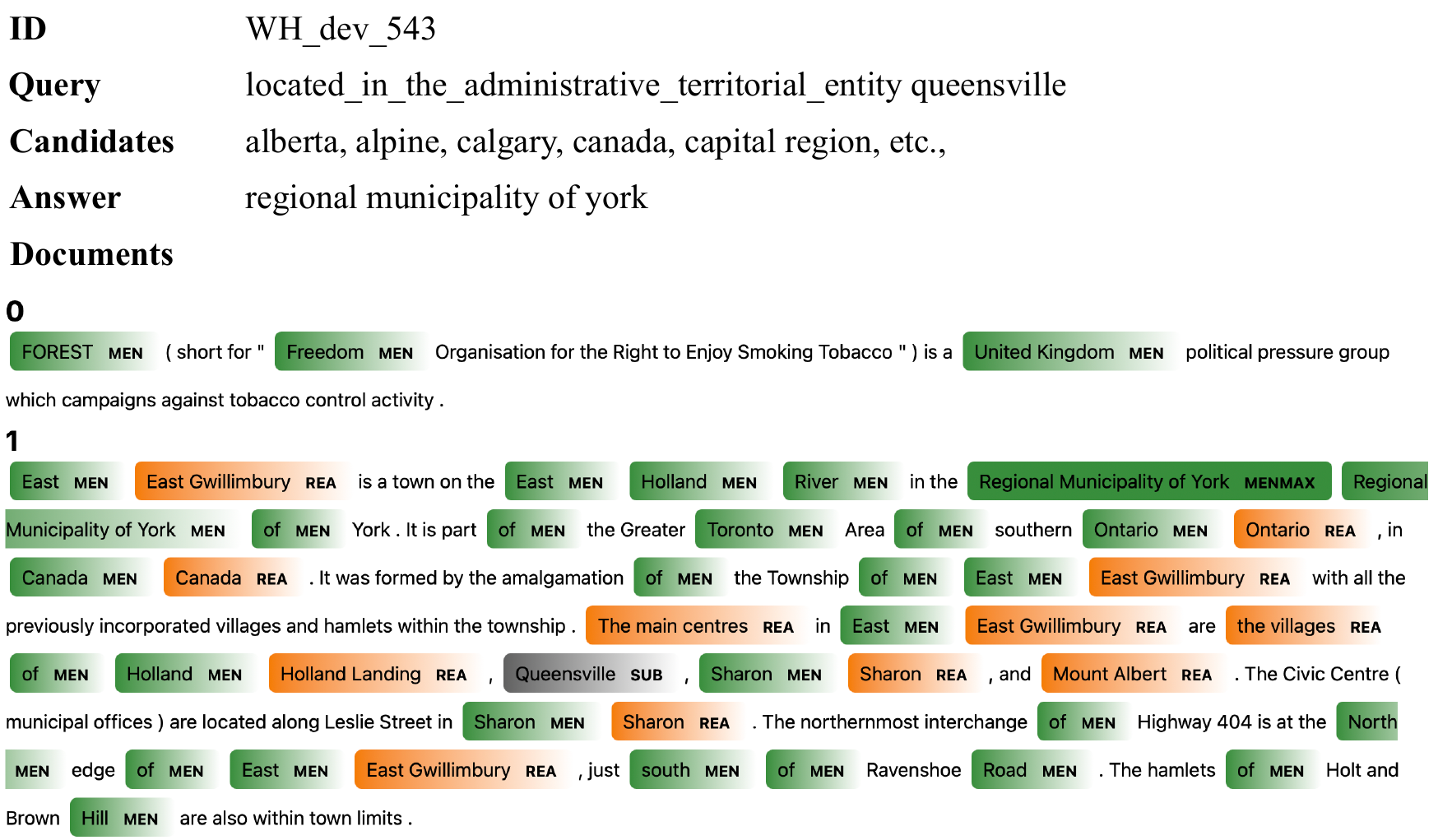}
    \caption{Generated HTML file of sample \# 543 in \WikiHop development set. The mark \textbf{MENMAX} means the final output of $\edit{\textbf{MLP}_{men}}$. For more details, please refer to \href{https://cluereader.github.io/WH\_dev\_543.html}{https://cluereader.github.io/WH\_dev\_543.html}.
    }
    \label{fig:html}
\end{figure}

\subsection{Visualization}

Compared to spectral GNN-based reading approaches, our proposed heterogeneous reasoning graph \crname is a non-spectral approach, which allows us to analyze how the nodes interact with each other in various relations and how the connections take effect between nodes.
We visualize the predictions in our heterogeneous reasoning graph on \WikiHop and \MedHop in Figures~\ref{fig:wikivis} and \ref{fig:medvis}, respectively. Different types of nodes are shown in different colors (subject nodes are gray, reasoning nodes are orange, mention nodes are green, candidate nodes are blue, and support nodes are red), and their edges, which reflect selections of node pairs, are shown in different thickness lines.
The thicker the edges, the more important they learn from training.
Considering that the answer determination should not only be inferred by the weight edges but also from the output layer projected from the representations of the nodes to $\mathbb{R}^{1\times2d}$ and accumulated score from $\edit{\mathscr{N}}_{can}$ and $\edit{\mathscr{N}}_{men}$, we use the transparency of the nodes to respond to the outputs: the darker the nodes, the higher the values output from the output layer.
Owing to the output values being quite different, some mention and candidate nodes are almost transparent.
The weight graph provides the evidence during reading and the analysis of DDI. It passes the messages according to the concept of \emph{grandmother cells} that not only one node becomes effective, but the cluster behind it plays a synergistic effect.
\edit{We learn more about our model through visualization.
For instance, the node transparency differentiation on \MedHop is significantly lower than \WikiHop, which indicates that the drug features are not sufficiently learned, leading to the convergence of node features and increased classification prediction difficulty.
This issue can be further addressed.}

To better understand the model predictions and contribute to further study, we generate HTML files of samples as shown in Figure~\ref{fig:html} and analyze whether the named entities contained in the max-score nodes can make sense from the perspective of a human answering after reading. Please refer to our website (\href{https://github.com/cluereader/cluereader.github.io}{https://github.com/cluereader/cluereader.github.io}) for more visualization samples in HTML files.

\section{Conclusion}\label{sec:5}
We present \crname, a heterogeneous graph attention network for multi-hop MRC, which is inspired by the concept of \emph{grandmother cells} from cognitive neuroscience.
The network contains several clue-reading paths from the subject of the question and ends with candidate entities. We use reasoning and mention nodes to complete the process and use support nodes to add supernumerary semantic information.
We apply our methodology on \qangaroo, a multi-hop MRC dataset, and the official evaluation supports the effectiveness of our model in open-domain QA and molecular biology domain.
Several potential issues could be further addressed, such as introducing intermediate supervision signals during the semi-supervised graph learning, the enhancement of using external knowledge, and dedicated word embedding methodology in the medical context, which are possible to improve the model performance in multi-hop MRC tasks.

\bibliography{manuscript}

\end{document}